\PassOptionsToPackage{hyperfootnotes=false}{hyperref}
\documentclass{SCYE}
\SubmissionMode
\usepackage[T1]{fontenc}
\usepackage[utf8]{inputenc}
\usepackage{array,multirow,makecell,longtable,xcolor}
\usepackage{setspace,flafter}
\usepackage{etoolbox,microtype}
\makeatletter
\expandafter\patchcmd\csname\string\title\endcsname
  {SCIENCE CHINA}{}{}{\PackageWarning{manuscript}{Banner text not found}}
\expandafter\patchcmd\csname\string\title\endcsname
  {Technological Sciences}{}{}{\PackageWarning{manuscript}{Banner subtitle not found}}
\let\relate@journal\@journal
\def\@journal{\phantom{\relate@journal}}
\makeatother
\hypersetup{
  pdftitle={Robust UAV-Satellite Geo-Localization with Degradation Benchmarking and Reliability-Guided Spatial Evidence Learning},
  pdfauthor={Haochen Jiang, Jialei Pan, Yuzhe Sun, Zhe Dong, Lecheng Ren, Yanfeng Gu, Tianzhu Liu}}
\providecommand{\bestres}[1]{\textcolor{red}{\textbf{#1}}}
\providecommand{\secondres}[1]{\textcolor{blue}{\textbf{#1}}}
\providecommand{\abgain}[1]{\textcolor{green!45!black}{\scriptsize #1}}
\providecommand{\metricgain}[2]{\begin{tabular}[c]{@{}c@{}}#1\\[-0.3mm]\abgain{#2}\end{tabular}}
\begin{document}
\twocolumn[{\vspace*{29.9168pt}
\ArticleType{Article}
\BeginPage{1}
\title{ReLATE: Reliable Evidence Learning for
	UAV–Satellite Geo-Localization}

\author[1]{Haochen JIANG}{}
\author[2]{Jialei PAN}{}
\author[1]{Yuzhe SUN}{}
\author[1]{Zhe DONG}{}
\author[3]{\protect\\Lecheng REN}{}
\author[1]{Yanfeng GU}{}
\author[1]{Tianzhu LIU}{tzliu@hit.edu.cn}
\AuthorMark{Jiang H C, et al.}
\AuthorCitation{Jiang H C, Pan J L, Sun Y Z, Dong Z, Ren L C, Gu Y F, Liu T Z}
\address[1]{School of Electronics and Information Engineering, Harbin Institute of Technology, Harbin {\rm 150001}, China}
\address[2]{National Key Laboratory of Radar Detection and Sensing,\\Nanjing Research Institute of Electronics Technology, Nanjing {\rm 210039}, China}
\address[3]{School of Electrical and Electronic Engineering, University of Manchester, Manchester {\rm M13 9PL}, United Kingdom}
\abstract{Unmanned aerial vehicle (UAV)-satellite cross-view geo-localization matches UAV images against satellite imagery and has achieved impressive accuracy on clean (non-degraded) image benchmarks. In real-world flights, however, UAV observations are frequently affected by adverse weather, illumination changes, platform motion, sensor noise, and compression, while the robustness of existing methods under such degradations remains largely unexamined. In this paper, we present UAVSat-Deg, a large-scale controlled degradation-synthesis
	benchmark for UAV--satellite geo-localization. It systematically
	simulates 27 single and compound acquisition-motivated degradations
	at three severity levels while preserving the original geospatial
	correspondences, and organizes them across bidirectional retrieval and
	multi-height UAV acquisition settings. The resulting benchmark contains more than 11.7 million pre-generated test images. Benchmarking representative methods under this protocol reveals substantial robustness gaps, particularly under severe and compound corruptions. To address this problem, we propose ReLATE, a Reliable Evidence Learning framework with Adaptive Token Evidence Regulation. ReLATE estimates a structure-smoothed reliability field over visual tokens to identify trustworthy spatial evidence and adaptively regulates its contribution to the cross-view descriptor, without relying on corrupted-image training, corruption labels, severity information, or prompts. Across both test sets and retrieval directions, ReLATE achieves the best average corrupted-test performance among the compared methods while maintaining competitive accuracy on clean images. The code and dataset
	will be available at \url{https://github.com/JHC626/RELATE}.}

\maketitle
}]
\AuthorMark{Jiang H C, et al.}
\begingroup
\renewcommand{\thefootnote}{}
\makeatletter
\long\def\@makefntext#1{\noindent#1}
\makeatother
\footnotetext{* Corresponding author: Tianzhu Liu (email: \href{mailto:tzliu@hit.edu.cn}{tzliu@hit.edu.cn}).\\Funding: National Natural Science Foundation of China, Grant Nos.~624B2051 (Special Program) and U25A6022.}
\endgroup
\setlength{\parindent}{10.8pt}
\renewcommand{\baselinestretch}{1.05}
\normalsize

\section{Introduction}
\label{sec:introduction}

UAV-satellite cross-view geo-localization aims to match UAV-view images with satellite-view images and retrieve their corresponding locations. It is a fundamental task for UAV navigation, remote sensing monitoring, emergency response, urban management, and low-altitude UAV perception. Typical settings include drone-to-satellite retrieval and satellite-to-drone retrieval, where a query image from one view is used to search for its corresponding target in the gallery of the other view. Unlike conventional image retrieval, UAV-satellite geo-localization must handle large viewpoint gaps, scale variations, altitude changes, spatial-layout discrepancies, and appearance shifts. Learning discriminative and view-consistent representations therefore remains a central challenge in this field~\cite{zheng2020university,zhu2023sues}.

\begin{figure}[tb]
    \centering
    \includegraphics[width=\columnwidth]{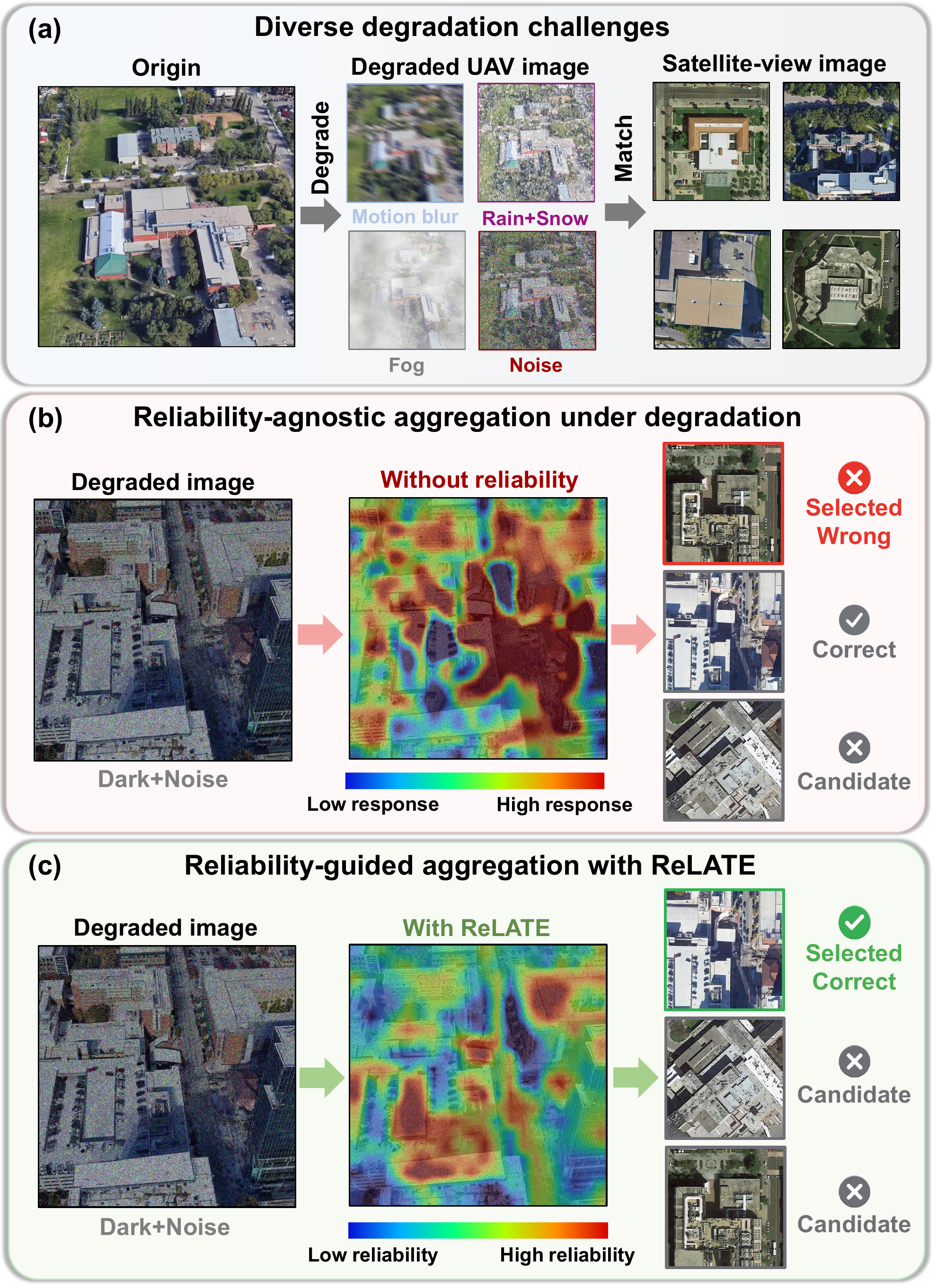}
    \caption{
		Motivation of reliability-guided evidence aggregation in ReLATE.
		(a) Diverse visual degradations, such as motion blur, rain, fog, and
		Gaussian noise, can corrupt UAV query images and increase the ambiguity
		of satellite-view retrieval.
		(b) Reliability-agnostic aggregation does not explicitly distinguish
		trustworthy local evidence from corruption-affected responses, which may
		allow unreliable evidence to enter the descriptor and cause an incorrect
		satellite match. The heatmap shows QDFL's normalized occlusion-based
		GT-matching contribution
		$E_{\mathrm{GT}}^{\mathrm{QDFL}}(p)$.
		(c) ReLATE estimates the reliability of local visual evidence,
		emphasizes high-reliability cues, and suppresses low-reliability
		responses, yielding the correct cross-view match in the illustrated
        example. The heatmap shows ReLATE's final structure-smoothed
        reliability field $\widetilde{\mathbf{R}}^{v}$. The two heatmaps depict
        different diagnostic quantities and are not on a common numerical scale.
	}
    \label{fig}
\end{figure}

Recent advances in deep metric learning, attention mechanisms, Transformer architectures, and visual foundation models have substantially improved UAV-satellite geo-localization on standard benchmarks~\cite{lin2015learning,hu2018cvm,shi2019spatial,yang2021cross,zhu2022transgeo,deuser2023sample4geo,dosovitskiy2020image,oquab2023dinov2}. Datasets such as University-1652 and SUES-200 have promoted drone-view and satellite-view retrieval by supporting studies on representation learning, local feature modeling, cross-view alignment, and retrieval optimization~\cite{zheng2020university,zhu2023sues}. However, existing evaluations are still largely conducted on clean (non-degraded) test sets, where both query and gallery images are assumed to have relatively ideal visual quality. Such clean-image evaluation is useful for measuring discriminative capability under standard benchmark conditions, but relying only on clean benchmarks makes it difficult to fully assess model stability and robustness under remote sensing imaging disturbances.

As shown in Fig.~\ref{fig}(a), in UAV and remote sensing applications, image quality is often affected by uncontrollable factors such as motion blur, defocus blur, sensor noise, compression artifacts, low illumination, adverse weather, haze, and reduced visibility. These degradations not only reduce visual quality, but may also corrupt key location-discriminative cues for cross-view matching, such as building contours, road structures, regional textures, vegetation distributions, and local object boundaries. UAV-satellite geo-localization already requires models to bridge geometric and appearance discrepancies between UAV and satellite views; when degraded observations further cause local evidence loss, structural ambiguity, or texture distortion, retrieval becomes substantially more challenging.

Recent studies have begun to investigate robustness in cross-view
geo-localization under visual corruptions, environmental changes, and
adverse weather, yet their evaluation settings and methodological
assumptions remain fragmented~
\cite{zhang2024benchmarking,wang2024multiple,feng2024multi,
	wen2026weatherprompt}. Despite these efforts, a unified protocol for
systematically evaluating the robustness of UAV-satellite
geo-localization across diverse visual degradations is still lacking.

To address this gap, we construct UAVSat-Deg, a systematic degraded
UAV-satellite cross-view geo-localization benchmark comprising
University-1652-Deg and SUES-200-Deg. UAVSat-Deg covers diverse
single and compound corruptions at multiple severity levels, supports
bidirectional drone-to-satellite and satellite-to-drone retrieval as
well as multi-height UAV acquisition, and follows a fixed image-only
clean-training (trained only on non-degraded images) corrupted-testing protocol.

To improve cross-view matching under degraded observations, we further propose ReLATE, i.e., Reliable Evidence Learning with Adaptive Token Evidence Regulation. ReLATE is a general reliable visual evidence learning framework rather than a model tailored to a specific degradation type. The key motivation is that different image regions and local visual responses contribute unequally to UAV-satellite matching reliability. Blur, noise, compression, and weather perturbations may produce unstable or misleading local evidence; indiscriminately aggregating such responses can weaken the location-discriminative capability of the final representation. 
As illustrated in Fig.~\ref{fig}(b) and Fig.~\ref{fig}(c), reliability-agnostic aggregation under degraded observations may introduce unreliable local responses into the descriptor and lead to incorrect retrieval, whereas ReLATE emphasizes high-reliability visual evidence and suppresses low-reliability responses for more robust cross-view matching.

ReLATE contains two complementary components.
Structure-Smoothed Reliability-Guided Evidence Learning (SRE) estimates
a spatially consistent reliability field and uses it to suppress unreliable
token responses while preserving stable structural cues.
Reliability-Adaptive Token Evidence Regulation (RATE) then aggregates
reliable token evidence and adaptively controls its contribution to the
final cross-view representation.

The main contributions of this paper are summarized as follows:
\begin{enumerate}
	\item We construct UAVSat-Deg, a systematic degraded UAV-satellite cross-view geo-localization benchmark, including University-1652-Deg and SUES-200-Deg. The benchmark covers multiple corruption families, severity levels, bidirectional retrieval tasks, and different UAV acquisition settings, and follows a reproducible image-only clean-training corrupted-testing protocol. Based on UAVSat-Deg, we systematically evaluate representative UAV-satellite geo-localization methods under corrupted observations, reveal their robustness gaps under weather, illumination changes, blur, noise, compression, and compound corruptions, and provide reproducible baselines and diagnostic evidence for future robust cross-view localization research.

	\item We propose ReLATE, a reliable visual evidence learning framework for robust UAV-satellite geo-localization. Its SRE module estimates a structure-smoothed reliability field that reduces the influence of unreliable visual responses while preserving stable spatial-structural cues, improving cross-view retrieval robustness under degraded remote sensing observations.
	
    \item We further introduce RATE, which converts the estimated reliability field into an active token evidence aggregation and descriptor regulation mechanism: reliability-weighted token evidence is distilled into a compact descriptor and adaptively injected into the query representations with input-dependent strength. This allows the final representation to exploit fine-grained spatial-structural cues when local evidence is trustworthy, and to avoid overemphasizing local token evidence when its reliability is limited. Extensive experiments show that ReLATE achieves stronger corrupted-test performance on both University-1652-Deg and SUES-200-Deg while maintaining competitive clean retrieval (retrieval without corruption) accuracy.

\end{enumerate}

The remainder of this paper is organized as follows. Section~\ref{sec:related_work} reviews related work on UAV-satellite cross-view geo-localization, degradation robustness evaluation, and generalizable cross-view representation learning. Section~\ref{sec:benchmark} introduces the construction and statistics of UAVSat-Deg. Section~\ref{sec:method} presents the proposed ReLATE framework. Section~\ref{sec:experiments} reports experimental settings, comparisons, and robustness analysis. Section~\ref{sec:ablation} presents the ablation studies, and Section~\ref{sec:conclusion} concludes this paper.

\section{Related Work}
\label{sec:related_work}

\subsection{UAV-Satellite Cross-View Geo-Localization}

UAV-satellite geo-localization is generally formulated as a bidirectional cross-view retrieval problem: a UAV image retrieves the corresponding satellite image, or a satellite query retrieves UAV observations of the same location. University-1652 established a multi-source benchmark for drone-view target localization and satellite-to-drone navigation~\cite{zheng2020university}, while SUES-200 introduced multi-height UAV acquisition and the associated scale and appearance variations~\cite{zhu2023sues}. Earlier ground-to-aerial studies developed wide-area localization, deep cross-view representations, orientation-aware alignment, and Transformer-based retrieval~\cite{workman2015wide,lin2015learning,hu2018cvm,shi2019spatial,yang2021cross,zhu2022transgeo}. Although their ground-view geometry differs from UAV imagery, these works provide important foundations for representation learning and cross-view alignment.

On UAV-satellite benchmarks, existing methods improve clean retrieval through complementary forms of spatial and metric modeling. LPN exploits part-level patterns and contextual information~\cite{wang2021each}; FSRA segments Transformer feature maps and aligns discriminative regions~\cite{dai2021transformer}; and Sample4Geo combines contrastive learning with hard-negative sampling~\cite{deuser2023sample4geo}. Other studies investigate practical UAV-satellite matching, view synthesis, keypoint-aware representation learning, multiple classifiers, domain alignment, contrastive attribute mining, self-adaptive feature extraction, counterfactual reasoning, and query-driven aggregation~\cite{ding2020practical,tian2021uav,lin2022joint,shen2023mccg,xia2024enhancing,wu2024camp,lin2024self,du2024ccr,hu2025query}. Recent extensions further consider field-of-view constraints, multiview scene matching, dynamic decorrelation, dense partition learning, Transformer aggregation, video-to-BEV transformation, language-guided navigation, and environment-independent feature enhancement~\cite{rodrigues2022global,sun2023f3,wang2024learning,chen2024sdpl,zhao2024transfg,ju2025video2bev,chu2024towards,zhao2025p2fcn}.

These advances have substantially improved retrieval on standard test sets, but most evaluations assume clean UAV and satellite images. Clean accuracy measures discriminability under the nominal data distribution; it does not show whether the same representation remains stable when local texture, visibility, illumination, or structural evidence is corrupted. This robustness gap motivates an evaluation protocol that varies corruption type and severity while keeping the training data and retrieval correspondences unchanged.

\begin{figure*}[tb]
    \centering
    \includegraphics[width=\textwidth]{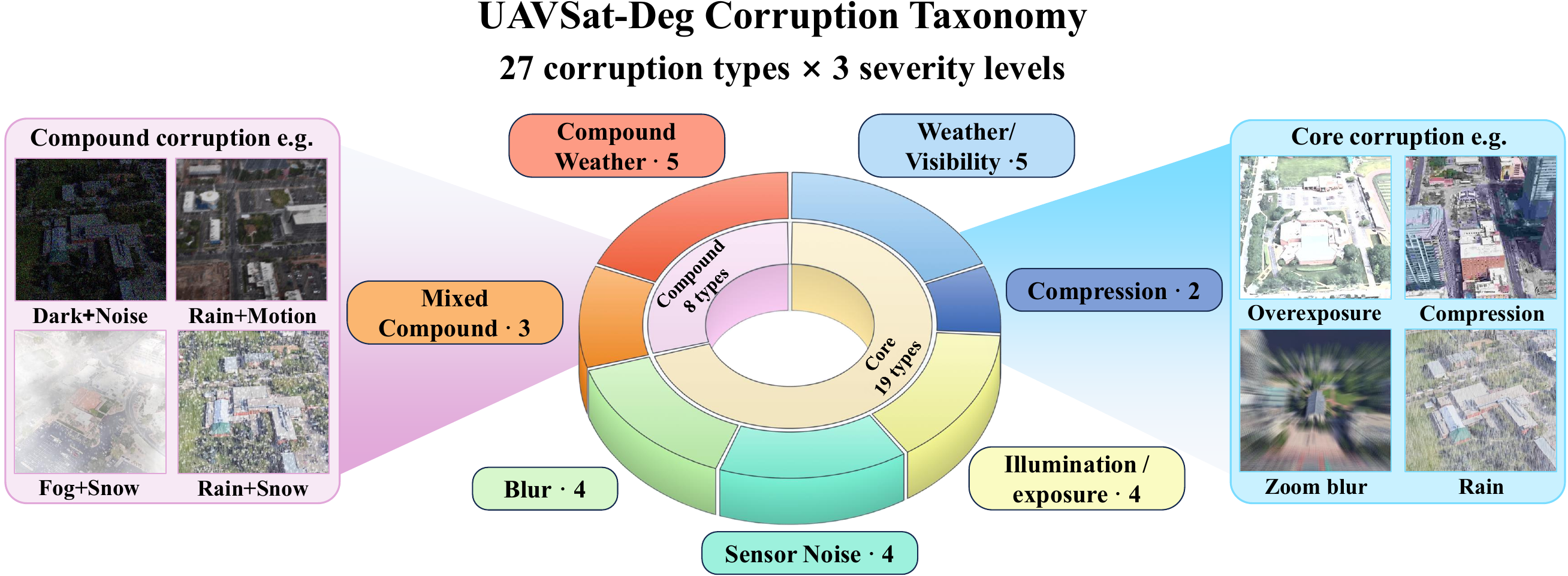}
    \caption{
		Overview of the UAVSat-Deg corruption taxonomy.
		The benchmark contains 27 corruption types at severity levels 1, 2, and 3: 19 core corruptions from five families and 8 compound corruptions.
		Representative compound and core examples are shown on the left and right, respectively.
	}
    \label{fig:dataset}
\end{figure*}

\subsection{Degraded and Robust Geo-Localization}

Corruption robustness has been systematically studied in generic recognition. ImageNet-C organizes noise, blur, weather, and digital artifacts into controlled severity levels, enabling standardized evaluation under distribution shift~\cite{hendrycks2019benchmarking}. A recent cross-view benchmark applies common corruptions to CVUSA and CVACT to analyze street-to-aerial retrieval robustness~\cite{zhang2024benchmarking}. These studies demonstrate that strong clean performance does not necessarily imply stable performance under visual perturbations. However, street-view matching differs from UAV-satellite localization in oblique viewpoint, altitude variation, and exposure to low-altitude motion, weather, and sensor disturbances.

Robust UAV geo-localization studies remain heterogeneous in both assumptions and protocols. Some treat weather or corruption as test-time perturbations, but cover only a few conditions, one dataset, or one retrieval direction. Others introduce degraded distributions during training. MuSe-Net models environment-induced shifts through multiple-environment style extraction and adaptive feature modulation~\cite{wang2024multiple}, whereas MCGF jointly optimizes diffusion-based restoration and geo-localization~\cite{feng2024multi}. Prompt-assisted and multimodal approaches adopt another setting: WeatherPrompt generates weather descriptions and learns text-guided representations with dynamic gating, drawing on vision-language priors such as CLIP~\cite{wen2026weatherprompt,radford2021learning}. Restoration, translation, degraded training samples, and text prompts
can improve robustness under their intended assumptions, but they
entangle the robustness of the underlying cross-view representation
with corruption-specific adaptation. In real-world deployment,
corruption types and severity levels are often unknown, mixed, or
previously unseen, so methods tailored to predefined degradations or
requiring explicit degradation descriptions may not transfer reliably
beyond the conditions covered during training.

We therefore adopt a fixed image-only clean-training corrupted-testing
protocol, in which all methods are trained on clean data and evaluated
on the same corrupted test sets without corruption labels, severity
information, prompts, or auxiliary modalities. This setting isolates
generalization to unseen degradations and enables a controlled
comparison of clean-trained cross-view representations.

\subsection{Generalizable Cross-View Representation Learning}
The core of cross-view localization is to keep observations of the same place close in feature space while separating geographically different locations. Existing approaches combine global descriptors, local or part-based regions, attention, Transformers, hard-negative sampling, and learnable queries~\cite{workman2015wide,lin2015learning,hu2018cvm,shi2019spatial,yang2021cross,zhu2022transgeo,wang2021each,dai2021transformer,deuser2023sample4geo,hu2025query}. Strong CNN and domain-generalization backbones~\cite{he2016deep,huang2017densely,pan2018two}, as well as Transformer and foundation-model representations~\cite{dosovitskiy2020image,liu2021swin,oquab2023dinov2}, provide rich semantic and structural features. Nevertheless, backbone strength alone does not guarantee degradation robustness.

Under corrupted observations, reliability is spatially nonuniform: roads, building outlines, and regional layouts may remain informative, whereas other tokens become unstable or misleading. Aggregation without explicit reliability modeling can therefore dilute the surviving geographic evidence. ReLATE addresses this issue without predicting a corruption category or inverting a degradation operator. Its SRE module estimates and spatially smooths token reliability, while RATE converts that reliability into adaptive token aggregation and descriptor regulation. The resulting representation retains the advantages of query-based cross-view modeling while explicitly controlling the contribution of degraded local evidence.

\section{Dataset Construction and Analysis}
\label{sec:benchmark}

\subsection{Dataset Construction}

To evaluate degradation robustness under controlled and reproducible conditions, we construct UAVSat-Deg from the test splits of University-1652~\cite{zheng2020university} and SUES-200~\cite{zhu2023sues}. The resulting University-1652-Deg and SUES-200-Deg retain the original training/test splits, location identities, class IDs, and retrieval correspondences; corruptions are applied only to UAV-side test images. University-1652-Deg covers the standard UAV-satellite setting, whereas SUES-200-Deg additionally preserves the H150, H200, H250, and H300 acquisition heights.

UAVSat-Deg obeys three principles~
\cite{hendrycks2019benchmarking,zhang2024benchmarking}.
First, all models are trained, selected, and tuned using only clean
data; degraded images are reserved exclusively for testing.
Second, every corrupted image is generated offline and fixed, so all
methods receive exactly the same inputs and the results are unaffected
by online random sampling. Third, the model receives only the image
itself. Corruption type, severity, generation parameters, weather
descriptions, prompts, and auxiliary modalities are never exposed.

These constraints reflect a practical deployment setting: future
corruption types, severity levels, and combinations cannot be
exhaustively enumerated during training, while explicit degradation
cues are typically unavailable at inference time. They therefore
separate degradation generalization from corruption-aware training or
test-time adaptation.

The taxonomy in Fig.~\ref{fig:dataset} comprises 19 core corruptions and 8 compound corruptions. Core corruptions represent individual imaging factors and are organized into blur, digital compression, illumination/exposure, sensor noise, and weather/visibility. Compound corruptions combine several factors in a fixed order and are divided into mixed and compound-weather groups. The taxonomy therefore covers both common single-factor disturbances and harder cases in which visibility, texture, and local structure deteriorate simultaneously.

For the weather-related subset, we follow the public WeatherPrompt-style synthesis procedure used in the original benchmark construction~\cite{wen2026weatherprompt}, but retain only the generated images and do not use weather descriptions, text embeddings, prompts, or multimodal reasoning. The remaining corruptions are produced with procedural image operators following the common-corruption paradigm, including sensor-noise, blur/optical, compression, and illumination transformations. Each type is rendered at severity levels 1, 2, and 3, representing mild, moderate, and severe degradation. The presets progressively change factor-specific parameters such as noise variance, blur extent, weather-particle density, opacity, illumination shift, and compression strength. Severity indices express comparable qualitative regimes across corruption types rather than identical physical magnitudes. For compound corruptions, component strengths are jointly balanced so that the scene remains partially recognizable even at severity 3.

Generation is retrieval-direction specific: UAV images are corrupted on the query side for drone-to-satellite (D2S) evaluation and on the gallery side for satellite-to-drone (S2D) evaluation, while satellite images remain clean in both directions. Because UAV images are acquired by moving onboard cameras, they are
more susceptible than relatively stable, pre-collected satellite
imagery to platform motion and vibration, rapidly varying illumination,
adverse weather, and sensor noise. The generated files are organized by dataset, direction, family, corruption type, and severity. SUES-200-Deg further retains all four flight heights, allowing degradation robustness to be analyzed jointly with scale and viewpoint changes.

\subsection{Dataset Statistics and Characteristics}

Table~\ref{tab:deg_corruption_taxonomy} lists the 19 core and
8 compound corruption types in UAVSat-Deg, each evaluated at severity
levels 1, 2, and 3.

\begin{table*}[tb]
	\centering
	\caption{Corruption taxonomy of UAVSat-Deg. Each corruption type is evaluated at severity levels 1, 2, and 3.}
	\label{tab:deg_corruption_taxonomy}
	\resizebox{\linewidth}{!}{%
\begin{tabular}{llll}
		\toprule
		Category & Family & Corruption Types & \#Types \\
		\midrule
		Core & Blur degradation &
		defocus\_blur, glass\_blur, motion\_blur, zoom\_blur & 4 \\
		Core & Digital compression &
		jpeg\_compression, pixelation & 2 \\
		Core & Illumination/exposure degradation &
		brightness, contrast, dark\_night, over\_exposure & 4 \\
		Core & Sensor noise &
		gaussian\_noise, impulse\_noise, shot\_noise, speckle\_noise & 4 \\
		Core & Weather/visibility degradation &
		fog, frost, rain, snow, wind & 5 \\
		\midrule
		Compound & Mixed compound degradation &
		dark\_noise, fog\_pixelation, rain\_motion & 3 \\
		Compound & Compound weather degradation &
		fog\_rain, fog\_snow, rain\_snow, fog\_rain\_snow, dark\_rain\_fog & 5 \\
		\bottomrule
	\end{tabular}%
	}
\end{table*}

For each retrieval direction, the core and compound groups produce \(19\times3=57\) and \(8\times3=24\) corrupted subsets, respectively. Thus, for each dataset, each direction contains 81 corrupted subsets and one clean subset. D2S and S2D together comprise 164 evaluation subsets per dataset, or 328 subsets across the two datasets. This organization supports macro-averaging by corruption type and separate analysis by family, severity, and retrieval direction.

University-1652-Deg uses 37,855 UAV queries in each D2S corruption-severity subset and 51,355 UAV gallery images in each S2D subset. Across 81 corrupted subsets, this yields 3,066,255 D2S and 4,159,755 S2D images, or 7,226,010 corrupted images in total. The benchmark additionally caches 90,862 clean reference images (the clean UAV sets and their satellite counterparts), bringing the complete University-1652-Deg cache to 7,316,872 images.

SUES-200-Deg caches 16,000 D2S UAV queries and 40,000
S2D UAV gallery images for each corruption--severity condition
across the four heights. Evaluation is performed separately at
each height, using 4,000 D2S queries and a 10,000-image S2D
gallery per height, and the four height-specific scores are
macro-averaged for the dataset-level results. It therefore contains 1,296,000 D2S and 3,240,000 S2D corrupted images, totaling 4,536,000. Across University-1652-Deg and SUES-200-Deg, UAVSat-Deg provides 11,762,010 corrupted test images. The core/compound and direction-wise breakdown is reported in Table~\ref{tab:deg_dataset_statistics}.

\begin{table*}[tb]
	\centering
	\caption{Statistics of University-1652-Deg and SUES-200-Deg. ``Images/subset'' denotes the number of corrupted images in one corruption-severity subset for the specified retrieval direction.}
	\label{tab:deg_dataset_statistics}
	\fontsize{9pt}{11pt}\selectfont
\setlength{\tabcolsep}{3pt}
\renewcommand{\arraystretch}{1.15}
\begin{tabular*}{\linewidth}{@{}l@{\extracolsep{\fill}}cccccc@{}}
		\toprule
		Dataset & Direction & \makecell{\#Corrupted\\Subsets} & \makecell{Images/\\Subset} & \makecell{Core\\Images} & \makecell{Compound\\Images} & \makecell{Total Corrupted\\Images} \\
		\midrule
		University-1652-Deg & D2S & 81 & 37,855 & 2,157,735 & 908,520 & 3,066,255 \\
		University-1652-Deg & S2D & 81 & 51,355 & 2,927,235 & 1,232,520 & 4,159,755 \\
		\midrule
		SUES-200-Deg & D2S & 81 & 16,000 & 912,000 & 384,000 & 1,296,000 \\
		SUES-200-Deg & S2D & 81 & 40,000 & 2,280,000 & 960,000 & 3,240,000 \\
		\bottomrule
	\end{tabular*}
\end{table*}

To quantitatively characterize the benchmark-wide degradation level, we
calculate the SNR of each corrupted UAV image relative to its corresponding
clean observation. After normalizing RGB intensities to $[0,1]$, the
clean-image energy is regarded as the signal power, while the mean-squared
difference between the corrupted and clean images is regarded as the
distortion power:
\begin{equation}
	\mathrm{SNR}
	=
	10\log_{10}
	\frac{\mathbb{E}[I^2]}
	{\mathbb{E}[(\widetilde{I}-I)^2]+\epsilon}.
\end{equation}
Across all corrupted UAV images in the two benchmarks, the median SNR is
approximately $9.05$ dB. This statistic provides an overall quantitative
description of the degradation strength, although for non-additive
corruptions it should be interpreted as image distortion relative to the clean
observation rather than physical sensor noise alone.

\section{Methodology}
\label{sec:method}

\subsection{Method Overview}

\begin{figure*}[tb]
	\centering
	\includegraphics[width=\textwidth]{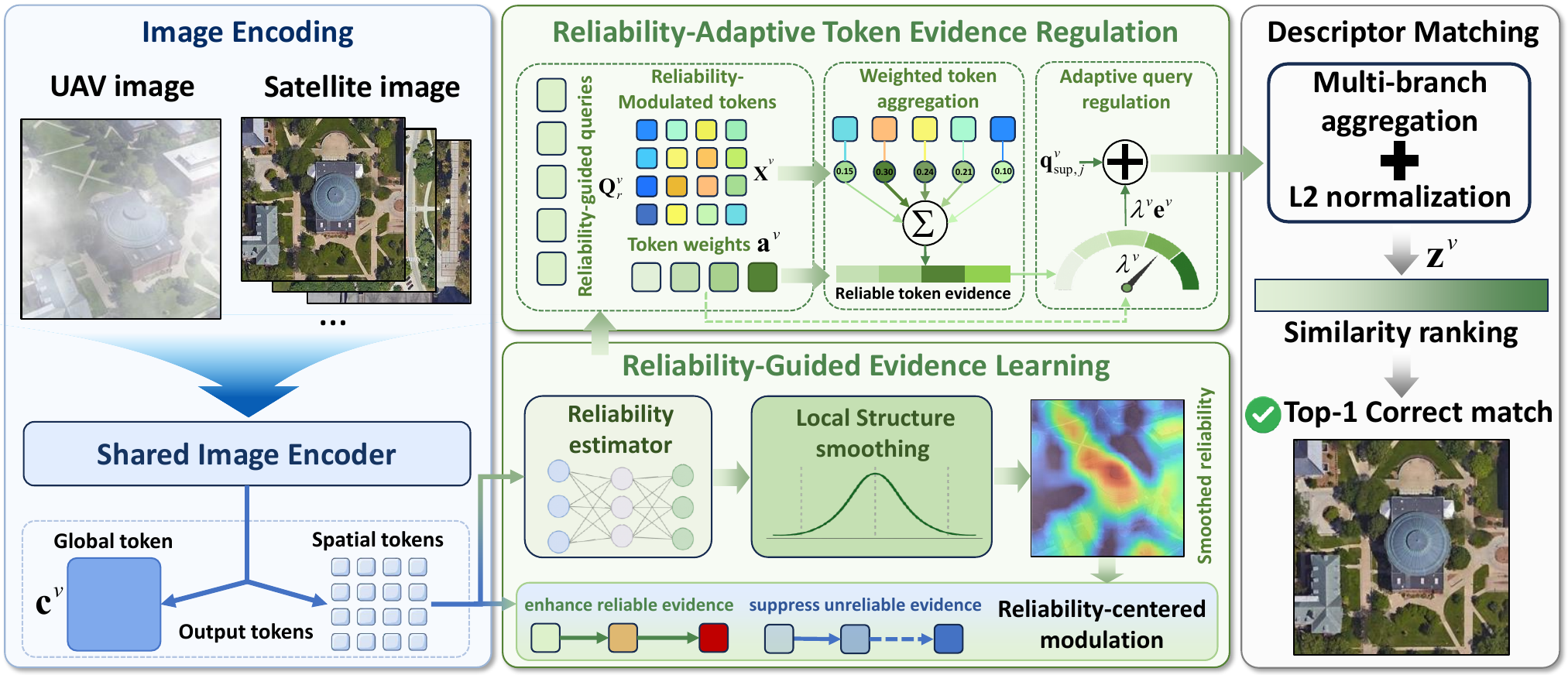}
	\caption{
		Overall framework of the proposed ReLATE.
		UAV-view and satellite-view images are independently encoded by a shared image encoder to obtain global and spatial tokens.
		The SRE module estimates and smooths token-level reliability, and modulates spatial tokens to enhance reliable evidence while suppressing unreliable responses.
		The RATE module aggregates reliable token evidence and adaptively regulates its contribution to the query representation.
		The $N_{\mathrm{sup}}$ regulated query representations, together with
		the CLS-token branch and the GeM-pooled spatial branch, are combined
		through multi-branch descriptor aggregation and L2-normalized to obtain
		the final descriptor $\mathbf{z}^v$, which is used for similarity ranking
		in UAV-satellite cross-view retrieval.
	}
	\label{fig:relate_framework}
\end{figure*}

As illustrated in Fig.~\ref{fig:relate_framework}, we propose ReLATE, namely Reliable Evidence Learning with Adaptive Token Evidence Regulation, for robust UAV-satellite cross-view geo-localization. ReLATE is not designed as a degradation-specific model; instead, it provides a general framework for reliable visual evidence learning. The central idea is that cross-view localization should not rely solely on stronger visual encoding, but should explicitly identify visual evidence that is stable and discriminative for location matching, and regulate how such evidence contributes to descriptor construction. Even in clean images, different regions contribute unequally to matching. Under corrupted observations, this imbalance becomes more pronounced, as blur, noise, compression artifacts, illumination changes, and weather perturbations may induce unstable or misleading local responses. Therefore, ReLATE builds a unified representation learning pipeline centered on reliable evidence discovery and reliable evidence utilization.

Given an input image $I^v$ from view $v \in \{d,s\}$, where $d$ and $s$ denote the drone and satellite views, respectively, ReLATE outputs a normalized descriptor for cross-view retrieval. A visual foundation encoder first extracts a global token and a set of spatial tokens:
\begin{equation}
	\mathbf{c}^{v}, \mathbf{X}^{v} = \Phi(I^{v}),
	\quad \mathbf{X}^{v}=\{\mathbf{x}^{v}_{i}\}_{i=1}^{N},
\end{equation}
where $\mathbf{c}^{v}$ denotes the global visual response and $\mathbf{X}^{v}$ denotes token-level spatial evidence. In our implementation, $\Phi(\cdot)$ is instantiated with a DINOv2 visual encoder based on Vision Transformer representations~\cite{dosovitskiy2020image,oquab2023dinov2}. ReLATE then performs reliable evidence learning over these token representations and produces the cross-view descriptor $\mathbf{z}^{v}$. During inference, the drone-view image and each satellite-view image are independently processed by the shared ReLATE framework. Let $\mathbf{z}^{d}$ and $\mathbf{z}^{s}$ denote the final $L_2$-normalized retrieval descriptors of the drone and satellite views, respectively. Their cross-view similarity is computed using the inner product:
\begin{equation}
	S(I^{d}, I^{s})
	=
	\left\langle
	\mathbf{z}^{d},
	\mathbf{z}^{s}
	\right\rangle.
\end{equation}
Candidate images are ranked according to the similarity scores for drone-to-satellite or satellite-to-drone retrieval.

The internal design of ReLATE follows a reliability perception-to-utilization pipeline. First, the SRE module estimates a spatially consistent reliability field over visual tokens and uses it to guide retrieval-oriented evidence learning. This module aims to identify trustworthy local responses, preserve stable spatial-structural cues such as road structures, building boundaries, and regional layouts, and reduce the influence of unstable local responses. Second, the RATE module converts the learned reliability field into an active token evidence regulation mechanism. It adaptively aggregates reliable token evidence and regulates its contribution to the final descriptor, enabling the cross-view representation to exploit fine-grained matching cues more effectively.

The above encoder-to-descriptor pipeline is instantiated on a
query-based cross-view representation substrate in line with
recent query-driven geo-localization~\cite{hu2025query}. Given the
spatial tokens $\mathbf{X}^{v}$, the substrate maintains
$K_q$ learnable query vectors and produces coarse evidence
queries
$\mathbf{Q}_{c}^{v}
=\{\mathbf{q}_{c,k}^{v}\}_{k=1}^{K_q}$
through its original query-encoding operations. SRE subsequently
refines these coarse queries by attending to the
reliability-modulated spatial tokens, yielding the
reliability-guided internal queries
$\mathbf{Q}_{r}^{v}
=\{\mathbf{q}_{r,k}^{v}\}_{k=1}^{K_q}$.
A learnable query-remapping layer then transforms the
$K_q$ internal queries into $N_{\mathrm{sup}}$ query-derived
supervised representations, which are further regulated by
RATE using the aggregated reliable token evidence. The
resulting $N_{\mathrm{sup}}$ regulated query branches,
together with one CLS-token branch and one GeM-pooled
spatial branch, form
$N_h=N_{\mathrm{sup}}+2$ final descriptor branches. The substrate does not explicitly model token reliability under degraded observations. ReLATE augments it with an explicit reliability layer: SRE estimates where local evidence can be trusted, and RATE regulates how much the trusted evidence contributes to the final descriptor.

\subsection{Structure-Smoothed Reliability-Guided Evidence Learning}

\begin{figure}[tb]
	\centering
	\includegraphics[width=\linewidth]{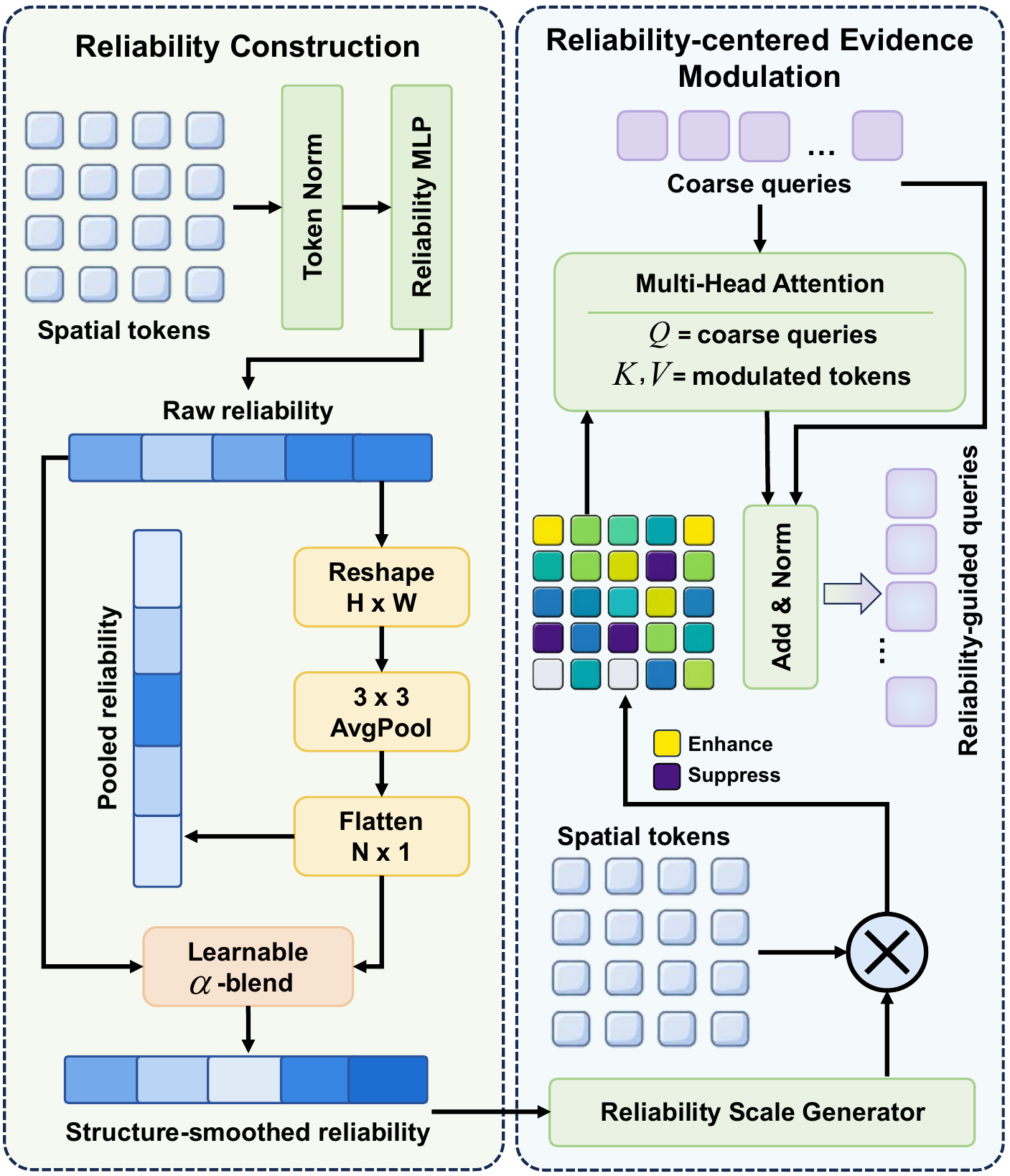}
	\caption{
		Illustration of structure-smoothed reliability-guided evidence learning.
		Given spatial tokens, ReLATE first estimates token-wise raw reliability through a reliability MLP and reshapes the reliability scores into a spatial map.
		A local average pooling operation and a learnable $\alpha$-blend are then used to obtain structure-smoothed reliability.
		The smoothed reliability is converted by a reliability scale generator into reliability-centered modulation scales, which enhance reliable token responses and suppress unreliable ones.
		Finally, coarse queries attend to the reliability-modulated spatial tokens through multi-head attention; the attention output is added back to the coarse queries and normalized to produce reliability-guided queries for subsequent descriptor construction.
	}
	\label{fig:innovation1}
\end{figure}

As illustrated in Fig.~\ref{fig:innovation1}, SRE estimates a spatially
consistent reliability field from visual tokens and uses it to modulate
token responses and refine retrieval queries.

Given the spatial token representations extracted by the visual encoder under view $v$,
\begin{equation}
	\mathbf{X}^{v}=\{\mathbf{x}^{v}_{i}\}_{i=1}^{N},
	\quad \mathbf{x}^{v}_{i}\in \mathbb{R}^{C},
\end{equation}
where $N=H\times W$ denotes the number of spatial tokens arranged on an $H\times W$ grid and $C$ denotes the channel dimension, we first estimate a soft reliability score for each token, providing a relative indication of how useful the corresponding local response is as cross-view matching evidence. Specifically, token reliability is predicted by a lightweight reliability estimator:
\begin{equation}
	r^{v}_{i}=\sigma\left(f_{r}(\mathbf{x}^{v}_{i})\right),
\end{equation}
where $f_{r}(\cdot)$ denotes a reliability mapping function composed of a normalization layer and a two-layer perceptron, $\sigma(\cdot)$ is the Sigmoid function, and $r^{v}_{i}\in(0,1)$. Here, reliability is an input-dependent latent utility score learned end-to-end for cross-view matching, rather than a calibrated probability of corruption or perceptual image quality. It is estimated solely from the visual response of the current token and does not require corruption labels, severity levels, or degradation parameters.

Pure token-wise reliability estimation can be sensitive to local noise and texture perturbations, especially in complex scenes or under compound corruptions, where independent token responses may yield fragmented or unstable reliability maps. In contrast, location-discriminative cues in UAV and remote sensing images usually exhibit spatial continuity. For example, roads, building groups, playgrounds, green areas, and regional boundaries typically appear as local structures rather than isolated tokens. Motivated by this observation, we perform structural smoothing on the reliability response. The token reliability scores $\mathbf{r}^{v}=[r^{v}_{1},\ldots,r^{v}_{N}]$ are reshaped into a spatial reliability map $\mathbf{R}^{v}\in\mathbb{R}^{H\times W}$, and a structural reliability response is obtained through a local smoothing operator $\mathcal{S}(\cdot)$:
\begin{equation}
	\label{eq:rel_smooth}
	\widetilde{\mathbf{R}}^{v}
	=
	\mathbf{R}^{v}
	+
	\alpha
	\left(
	\mathcal{S}(\mathbf{R}^{v})-\mathbf{R}^{v}
	\right),
\end{equation}
where $\mathcal{S}(\cdot)$ is instantiated as $3\times3$ average pooling with stride 1 and padding 1, and $\alpha\in(0,1)$ is a learnable smoothing coefficient constrained by a Sigmoid parameterization, so that $\widetilde{\mathbf{R}}^{v}$ is a convex combination of the raw and locally pooled reliability. This design allows reliability estimation to incorporate local neighborhood structures rather than relying on isolated token responses: the local averaging encourages spatial consistency of the reliability field, while the residual blend in Eq.~\eqref{eq:rel_smooth} retains the original token-wise reliability variation, yielding a less fragmented and more stable reliability field consistent with the spatial continuity of remote sensing imagery. The smoothed reliability map is then flattened back into $\widetilde{\mathbf{r}}^{v}=\{\widetilde{r}^{v}_{i}\}_{i=1}^{N}$ for subsequent token-wise modulation and evidence aggregation.

After obtaining the structure-smoothed reliability field, we use it to modulate spatial tokens. To reduce uniform amplification or attenuation across the token set, the reliability scores are first mean-centered:
\begin{equation}
	\Delta r^{v}_{i}
	=
	\widetilde{r}^{v}_{i}
	-
	\frac{1}{N}
	\sum_{j=1}^{N}
	\widetilde{r}^{v}_{j}.
\end{equation}
A token-level modulation factor is then generated from the centered reliability:
\begin{equation}
	m^{v}_{i}
	=
	\mathrm{clip}
	\left(
    1+\gamma \Delta r^{v}_{i},
    m_{\min},
    m_{\max}
    \right),
\qquad \gamma>0,
\end{equation}
where $\gamma>0$ is a learnable modulation strength kept positive
through a Softplus parameterization, i.e.,
$\gamma=\operatorname{softplus}(\eta_\gamma)$ for a learnable scalar
$\eta_\gamma$, and $0<m_{\min}<1<m_{\max}$ ensures positive and
bounded feature scaling. The reliability-modulated token representation is defined as
\begin{equation}
	\widehat{\mathbf{x}}^{v}_{i}
	=
	m^{v}_{i}\mathbf{x}^{v}_{i}.
\end{equation}
In this way, tokens with above-average reliability are enhanced, whereas tokens with below-average reliability are suppressed. Because the modulation is based on centered reliability, it does not simply amplify the whole feature map, but emphasizes the relative credibility of local evidence and guides the model toward more location-discriminative spatial-structural cues.

During retrieval-oriented evidence learning, the reliability-modulated tokens are used to refine the view-specific query representations employed for subsequent cross-view matching. Recall the coarse evidence queries $\mathbf{Q}^{v}_{c}$ from the representation substrate, which aggregate retrieval-relevant contextual information from the spatial tokens. We use the reliability-modulated token representations $\widehat{\mathbf{X}}^{v}$ as key and value, and obtain reliability-guided query representations through cross-attention:
\begin{equation}
	\mathbf{Q}^{v}_{r}
	=
	\mathrm{LN}
	\left(
	\mathbf{Q}^{v}_{c}
	+
	\mathrm{MHA}
	\left(
	\mathrm{LN}(\mathbf{Q}^{v}_{c}),
	\widehat{\mathbf{X}}^{v},
	\widehat{\mathbf{X}}^{v}
	\right)
	\right),
\end{equation}
where $\mathrm{MHA}(\cdot)$ denotes multi-head attention~\cite{vaswani2017attention} and $\mathrm{LN}(\cdot)$ denotes layer normalization. Through this reliability-guided attention, the model reduces the influence of less reliable tokens during the
query--token interaction. Instead, it is encouraged to focus on reliable structural cues and reduce the interference of unstable local responses in the final descriptor.

\subsection{Reliability-Adaptive Token Evidence Regulation}

The structure-smoothed reliability-guided evidence learning module estimates a reliability field in the visual token space and uses it to guide retrieval-oriented evidence learning. While this reliability-guided refinement reduces the influence of unreliable local responses during feature learning, robust descriptor construction further requires an explicit mechanism to determine how reliable token-level evidence should contribute to the final representation. UAV-satellite cross-view geo-localization is a fine-grained retrieval task. Its matching performance depends not only on global scene context but also on local spatial-structural cues, such as road layouts, building boundaries, regional contours, and texture patterns. These cues are often encoded by local tokens. Relying only on a globally aggregated descriptor may weaken fine-grained matching information, whereas indiscriminately aggregating all tokens may introduce unstable local responses into the final representation. To this end, we propose Reliability-Adaptive Token Evidence Regulation, which converts the reliability field into an active token evidence aggregation and descriptor regulation mechanism.

Given the structure-smoothed reliability score $\widetilde{r}^{v}_{i}$ and the corresponding reliability-modulated token representation $\widehat{\mathbf{x}}^{v}_{i}$ produced by SRE, we assign adaptive evidence weights according to the reliability distribution using a concentration-controlled softmax:
\begin{equation}
	a^{v}_{i}
	=
	\frac{
		\exp(\tau \widetilde{r}^{v}_{i})
	}{
		\sum_{j=1}^{N}
		\exp(\tau \widetilde{r}^{v}_{j})
	},
\end{equation}
where $\tau>0$ controls the concentration of the weight distribution. Because softmax depends on relative logit differences, tokens with higher reliability receive larger evidence weights, whereas less reliable tokens are suppressed. Unlike average pooling or fixed-weight aggregation, the weights are determined by the current input, allowing the model to select trustworthy local cues across images, views, and imaging conditions.

Based on the token evidence weights, local token representations are aggregated into a reliable token evidence descriptor:
\begin{equation}
	\mathbf{e}^{v}
	=
	\mathrm{LN}
	\left(
	\sum_{i=1}^{N}
	a^{v}_{i}
	\widehat{\mathbf{x}}^{v}_{i}
	\right),
\end{equation}
where $\mathrm{LN}(\cdot)$ denotes layer normalization. This descriptor serves as a compact summary of reliable local evidence extracted from the current image. Compared with directly averaging all tokens, this aggregation emphasizes stable and location-discriminative local structural cues, thereby providing complementary information for the final cross-view descriptor. Note that the modulation factor $m^{v}_{i}$ in SRE and the aggregation weight $a^{v}_{i}$ here serve different purposes: the former rescales token responses before the query--token interaction, whereas the latter determines the contribution of each token during explicit token evidence aggregation. The two operations therefore act at different stages of the pipeline and are complementary rather than redundant.

Furthermore, the contribution of token evidence to the final descriptor should be input-dependent. Across different input images, views, and UAV acquisition conditions, the reliability and necessity of local token evidence may vary. When the global structure is clear, token evidence can provide fine-grained complementary cues; when local textures are severely perturbed, overemphasizing token evidence may introduce additional noise. Therefore, we introduce a reliability-adaptive regulation coefficient $\lambda^{v}$ to control the contribution of reliable token evidence to the final cross-view representation:
\begin{equation}
	\mathbf{s}^{v}
	=
	\Psi(\widetilde{\mathbf{R}}^{v})
	=
	\big[
	\operatorname{mean}(\widetilde{\mathbf{R}}^{v}),\,
	\operatorname{std}(\widetilde{\mathbf{R}}^{v})
	\big],
\end{equation}
\begin{equation}
	\lambda^{v}
	=
	\sigma\!\left(
	\mathrm{MLP}_{\lambda}(\mathbf{s}^{v})
	\right)
	\in(0,1),
\end{equation}
where $\Psi(\cdot)$ summarizes the structure-smoothed reliability field $\widetilde{\mathbf{R}}^{v}$ by its global distribution statistics, $\mathrm{MLP}_{\lambda}(\cdot)$ denotes a lightweight two-layer perceptron, and the Sigmoid function $\sigma(\cdot)$ bounds the regulation coefficient to $\lambda^{v}\in(0,1)$. In this way, the integration strength of token evidence is adaptively determined by the overall level and dispersion of the reliability distribution of the current input, enabling the model to flexibly regulate the influence of local evidence under different reliability states, views, and acquisition conditions.

Recall the reliability-guided internal query representations
$\mathbf{Q}^{v}_{r}
=
\{\mathbf{q}^{v}_{r,k}\}_{k=1}^{K_q}$
produced by the SRE module, where $K_q$ denotes the number of
internal queries. Before branch-level descriptor construction, a
learnable query-remapping function $\mathcal{R}_{q}(\cdot)$ transforms
the $K_q$ internal queries along the query dimension into
$N_{\mathrm{sup}}$ query-derived supervised representations. The
reliable token evidence descriptor is then injected into each remapped
query representation:
\begin{subequations}
	\label{eq:query_regulation}
	\begin{gather}
		\mathbf{Q}^{v}_{\mathrm{sup}}
		=
		\mathcal{R}_{q}
		\left(
		\mathbf{Q}^{v}_{r}
		\right),
		\label{eq:query_remapping}
		\\
		\mathbf{q}^{v}_{\mathrm{reg},j}
		=
		\mathbf{q}^{v}_{\mathrm{sup},j}
		+
		\lambda^{v}\mathbf{e}^{v},
		\qquad
		j=1,\ldots,N_{\mathrm{sup}}.
		\label{eq:query_evidence_injection}
	\end{gather}
\end{subequations}
Here, 
$\mathbf{Q}^{v}_{\mathrm{sup}}
\in
\mathbb{R}^{N_{\mathrm{sup}}\times C}$,
and $\mathbf{q}^{v}_{\mathrm{sup},j}$ denotes its $j$-th row.
The function
$\mathcal{R}_{q}:
\mathbb{R}^{K_q\times C}
\rightarrow
\mathbb{R}^{N_{\mathrm{sup}}\times C}$
performs learnable linear remapping along the query dimension.
Therefore, the reliable token evidence is injected into the
query-derived supervised representations rather than treating all
internal queries as independent descriptor branches.

In addition to the regulated query branches, the representation
substrate retains a CLS-token branch and a GeM-pooled spatial branch.
Let
\begin{equation}
	\mathbf{g}^{v}
	=
	\operatorname{GeM}
	\left(
	\mathbf{X}^{v}_{\mathrm{f}}
	\right),
\end{equation}
where $\mathbf{X}^{v}_{\mathrm{f}}$ denotes the refined spatial feature
map retained by the query-based representation substrate and used as
the input to the GeM-pooled spatial branch. The branch-level descriptors are
then defined as
\begin{subequations}
	\label{eq:branch_descriptors}
	\begin{gather}
		\mathbf{u}^{v}_{j}
		=
		\mathcal{P}_{j}
		\left(
		\mathbf{q}^{v}_{\mathrm{reg},j}
		\right),
		\quad
		j=1,\ldots,N_{\mathrm{sup}},
		\label{eq:query_branch_descriptor}
		\\
		\mathbf{u}^{v}_{\mathrm{cls}}
		=
		\mathcal{P}_{\mathrm{cls}}
		\left(
		\mathbf{c}^{v}
		\right),
		\label{eq:cls_branch_descriptor}
		\\
		\mathbf{u}^{v}_{\mathrm{gem}}
		=
		\mathcal{P}_{\mathrm{gem}}
		\left(
		\mathbf{g}^{v}
		\right).
		\label{eq:gem_branch_descriptor}
	\end{gather}
\end{subequations}
The first $N_{\mathrm{sup}}$ descriptors are query-derived
supervised branches, while
$\mathbf{u}^{v}_{\mathrm{cls}}$
and
$\mathbf{u}^{v}_{\mathrm{gem}}$
denote the CLS-token and GeM-pooled branches, respectively.
Therefore, the total number of descriptor branches is
$N_h=N_{\mathrm{sup}}+2$.

The final cross-view retrieval descriptor concatenates all branch
descriptors and applies $L_2$ normalization:
\begin{equation}
	\begin{aligned}
		\mathbf{z}^{v}
		=
		\operatorname{Norm}\Big(
		\operatorname{Concat}\big[
		&\mathbf{u}_{1}^{v},
		\ldots,
		\mathbf{u}_{N_{\mathrm{sup}}}^{v},\\[-1mm]
		&\mathbf{u}_{\mathrm{cls}}^{v},
		\mathbf{u}_{\mathrm{gem}}^{v}
		\big]\Big).
	\end{aligned}
\end{equation}
The descriptor $\mathbf{z}^{v}$ is used for cross-view
retrieval, and the complete set of
$N_h=N_{\mathrm{sup}}+2$ branch descriptors additionally
serves as the input to the location classification heads
described in Section~\ref{sec:training_objective}.

\subsection{Training Objective}
\label{sec:training_objective}

ReLATE does not rely on any explicit reliability annotation or degradation label. The reliability field is learned implicitly, driven entirely by the retrieval objective. Following the training recipe of the query-based
substrate~\cite{hu2025query}, each view produces
$N_h=N_{\mathrm{sup}}+2$ branch-level location predictions,
including $N_{\mathrm{sup}}$ regulated query branches, one
CLS-token branch, and one GeM-pooled spatial branch, together
with a final cross-view descriptor. The overall training
objective combines three complementary terms:
\begin{equation}
	\mathcal{L}
	=
	\mathcal{L}_{\mathrm{cls}}
	+
	\mathcal{L}_{\mathrm{MS}}
	+
	\mathcal{L}_{\mathrm{cv}}.
\end{equation}
The three loss terms are combined with equal weights, and no additional loss-balancing hyperparameters are introduced.
\noindent\textbf{Location classification loss.}
For each view, the $N_h$ classification branches are supervised by the shared location label $y$. The $h$-th branch feeds its branch descriptor into a linear classifier to produce the location logits, i.e., $\mathbf{o}^{v}_{h}=\mathbf{W}_{h}\mathbf{u}^{v}_{h}+\mathbf{b}_{h}$, where $\mathbf{W}_{h}$ and $\mathbf{b}_{h}$ are shared between the two views; we write $\mathbf{o}^{s}_{h}$ and $\mathbf{o}^{d}_{h}$ for the satellite and drone views, respectively. The classification loss averages the cross-entropy over all branches of each view and sums the two views:
\begin{equation}
	\mathcal{L}_{\mathrm{cls}}
	=
    \frac{1}{N_h}\sum_{h=1}^{N_h}
	\left[
	\mathrm{CE}(\mathbf{o}^{s}_{h}, y)
	+
	\mathrm{CE}(\mathbf{o}^{d}_{h}, y)
	\right].
\end{equation}

\noindent\textbf{Cross-view Multi-Similarity loss.}
To shape a discriminative retrieval embedding, we impose a Multi-Similarity loss $\ell_{\mathrm{MS}}(\cdot)$~\cite{wang2019multi} on the final cross-view descriptor. The satellite and UAV descriptors within a batch are concatenated and assigned the same location labels, so that descriptors of the same location form positive pairs and descriptors of different locations form negative pairs. Denote the concatenated descriptor and label sets as
\begin{equation}
	\mathbf{F}
	=
	\operatorname{Concat}(\mathbf{F}^{s}, \mathbf{F}^{d}),
	\qquad
	\mathbf{Y}
	=
	\operatorname{Concat}(\mathbf{y}, \mathbf{y}),
\end{equation}
where $\mathbf{F}^{s}=\{\mathbf{z}^{s}_{b}\}_{b=1}^{B}$ and $\mathbf{F}^{d}=\{\mathbf{z}^{d}_{b}\}_{b=1}^{B}$ denote the satellite and drone descriptor sets in a batch of size $B$, and $\mathbf{y}$ denotes the shared location labels. A Multi-Similarity miner with threshold $\epsilon_{\mathrm{MS}}$ first selects informative pairs from the joint cross-view descriptor set, and the metric loss is defined as
\begin{equation}
	\mathcal{L}_{\mathrm{MS}}
	=
	\ell_{\mathrm{MS}}(\mathbf{F}, \mathbf{Y}).
\end{equation}
We set $\alpha_{\mathrm{MS}}=5$, $\beta_{\mathrm{MS}}=100$, base similarity $b_{\mathrm{MS}}=0.1$, and mining threshold $\epsilon_{\mathrm{MS}}=0.4$.

\noindent\textbf{Cross-view prediction-consistency loss.}
To encourage the two views to produce consistent location predictions, we align their per-branch class distributions toward a shared mixture target. For the $h$-th branch, let $\mathbf{p}^{s}_{h}$ and $\mathbf{p}^{d}_{h}$ be the softmax class distributions of the two views, and let $\mathbf{m}_{h}=\mathrm{sg}\!\left(\tfrac{1}{2}(\mathbf{p}^{s}_{h}+\mathbf{p}^{d}_{h})\right)$ be the averaged mixture distribution, where $\mathrm{sg}(\cdot)$ denotes the stop-gradient operation. The consistency loss is a symmetric mixture-to-view Kullback--Leibler term:
\begin{equation}
	\mathcal{L}_{\mathrm{cv}}
	=
    \frac{1}{2N_h}\sum_{h=1}^{N_h}
	\left[
	D_{\mathrm{KL}}(\mathbf{m}_{h} \,\|\, \mathbf{p}^{s}_{h})
	+
	D_{\mathrm{KL}}(\mathbf{m}_{h} \,\|\, \mathbf{p}^{d}_{h})
	\right].
\end{equation}
Rather than directly minimizing a bidirectional Kullback--Leibler divergence between the two view predictions, this formulation aligns both of them to a shared stop-gradient mixture target. This term constrains only the cross-view location prediction distributions and introduces no additional reliability or degradation supervision, so the reliability modeling in ReLATE remains an intermediate representation optimized end-to-end by the retrieval objective.
\section{Experiments}
\label{sec:experiments}

\begin{table}[tb]
	\centering
	\caption{Severity-wise corrupted-test comparison of Ours and QDFL.
		Entries are R@1/AP averaged over all corruption types at each severity level.
		For SUES-200-Deg, results are further macro-averaged over the four UAV heights.
		D$\rightarrow$S and S$\rightarrow$D denote Drone$\rightarrow$Satellite
		and Satellite$\rightarrow$Drone, respectively.}
	\label{tab:severity_level_summary}
	
	\fontsize{8pt}{9pt}\selectfont
	\setlength{\tabcolsep}{0.9pt}
	\renewcommand{\arraystretch}{1.08}
	\setlength{\aboverulesep}{0.30ex}
	\setlength{\belowrulesep}{0.30ex}
	
	\begin{tabular}{@{}llcccc@{}}
		\toprule
		\shortstack[l]{Dataset /\\Backbone}
		& Direction
		& Method
		& Sev. 1
		& Sev. 2
		& Sev. 3 \\
		\midrule
		
		\multirow{4}{*}{
			\shortstack[l]{
				Univ.-1652-Deg\\
				DINOv2-B/14
			}
		}
		& \multirow{2}{*}{D$\rightarrow$S}
		& QDFL
		& 85.36/87.30
		& 66.33/69.77
		& 44.50/48.59 \\
		
		&
		& \textbf{Ours}
		& \textcolor{red}{\textbf{87.76/89.48}}
		& \textcolor{red}{\textbf{71.23/74.43}}
		& \textcolor{red}{\textbf{50.25/54.29}} \\
		
		\cmidrule(lr){2-3}
		\cmidrule(lr){4-6}
		
		& \multirow{2}{*}{S$\rightarrow$D}
		& QDFL
		& 94.35/86.46
		& 90.41/74.86
		& 81.08/57.72 \\
		
		&
		& \textbf{Ours}
		& \textcolor{red}{\textbf{94.85/87.63}}
		& \textcolor{red}{\textbf{91.41/77.10}}
		& \textcolor{red}{\textbf{83.16/60.87}} \\
		
		\midrule
		
		\multirow{4}{*}{
			\shortstack[l]{
				SUES-200-Deg\\
				DINOv2-B/14
			}
		}
		& \multirow{2}{*}{D$\rightarrow$S}
		& QDFL
		& 88.47/92.37
		& 71.95/79.64
		& 50.99/61.19 \\
		
		&
		& \textbf{Ours}
		& \textcolor{red}{\textbf{90.65/93.98}}
		& \textcolor{red}{\textbf{75.94/82.79}}
		& \textcolor{red}{\textbf{54.74/64.58}} \\
		
		\cmidrule(lr){2-3}
		\cmidrule(lr){4-6}
		
		& \multirow{2}{*}{S$\rightarrow$D}
		& QDFL
		& 97.04/91.12
		& 90.65/79.23
		& 78.31/61.82 \\
		
		&
		& \textbf{Ours}
		& \textcolor{red}{\textbf{98.04/92.33}}
		& \textcolor{red}{\textbf{92.89/80.84}}
		& \textcolor{red}{\textbf{80.14/62.86}} \\
		
		\bottomrule
	\end{tabular}
\end{table}

\begin{table}[tb]
	\centering
	\caption{
		Rank-consistency summary across corrupted evaluation conditions.
		Each dataset contains 54 conditions, corresponding to two retrieval
		directions and 27 corruption types. For each condition, methods are
		ranked according to R@1. Ranks 1, 2, 3, 4, and 5 or lower receive
		5, 4, 3, 2, and 1 points, respectively, and Rating denotes the
		average score over all 54 conditions. Best, Top-2, and Top-3 report
		cumulative condition counts. Clean and All-27 entries are excluded
		to avoid double counting. Tied results share the same rank.
	}
	\label{tab:rank_consistency_summary}
	
	\fontsize{8pt}{9.5pt}\selectfont
	\renewcommand{\arraystretch}{1.06}
	\setlength{\tabcolsep}{4.0pt}
	
	\begin{minipage}[t]{\linewidth}
		\centering
		\textbf{(a) University-1652-Deg}
		
		\vspace{0.6mm}
		
		\begin{tabular}{lcccc}
			\toprule
			Method & Best & Top-2 & Top-3 & Rating \\
			\midrule
			\textbf{Ours}
			& \textbf{39/54}
			& \textbf{50/54}
			& \textbf{54/54}
			& \textbf{4.65/5} \\
			
			QDFL       & 1/54  & 39/54 & 49/54 & 3.65/5 \\
			DAC        & 11/54 & 15/54 & 41/54 & 3.15/5 \\
			CAMP       & 0/54  & 1/54  & 4/54  & 1.61/5 \\
			MuSe-Net   & 1/54  & 1/54  & 8/54  & 1.46/5 \\
			Sample4Geo & 2/54  & 2/54  & 2/54  & 1.20/5 \\
			MCCG       & 0/54  & 0/54  & 2/54  & 1.19/5 \\
			FSRA       & 0/54  & 0/54  & 2/54  & 1.09/5 \\
			MEAN       & 0/54  & 0/54  & 0/54  & 1.00/5 \\
			\bottomrule
		\end{tabular}
	\end{minipage}
	\par\vspace{2mm}
	\begin{minipage}[t]{\linewidth}
		\centering
		\textbf{(b) SUES-200-Deg}
		
		\vspace{0.6mm}
		
		\begin{tabular}{lcccc}
			\toprule
			Method & Best & Top-2 & Top-3 & Rating \\
			\midrule
			\textbf{Ours}
			& \textbf{39/54}
			& \textbf{51/54}
			& \textbf{54/54}
			& \textbf{4.67/5} \\
			
			QDFL       & 3/54 & 35/54 & 45/54 & 3.52/5 \\
			CAMP       & 5/54 & 13/54 & 29/54 & 2.70/5 \\
			Sample4Geo & 7/54 & 12/54 & 30/54 & 2.70/5 \\
			MCCG       & 0/54 & 0/54  & 5/54  & 1.31/5 \\
			CCR        & 0/54 & 0/54  & 0/54  & 1.11/5 \\
			DAC        & 0/54 & 0/54  & 0/54  & 1.06/5 \\
			FSRA       & 0/54 & 0/54  & 0/54  & 1.00/5 \\
			MEAN       & 0/54 & 0/54  & 0/54  & 1.00/5 \\
			\bottomrule
		\end{tabular}
	\end{minipage}
	
	\vspace{-1mm}
\end{table}

\begin{table*}[tb]
	\centering
	\caption{
		Comparison on University-1652-Deg for Drone $\rightarrow$ Satellite retrieval.
		Entries are R@1/AP averaged over severity levels.
		All-27 denotes the average over all 27 corrupted types and excludes Clean.
		Best and second-best results are marked in
		\textcolor{red}{red} and \textcolor{blue}{blue}, respectively.
	}
	\label{tab:u1652_drone_to_sat_all}
	
	\renewcommand{\baselinestretch}{1}
	\fontsize{7pt}{8pt}\selectfont
	\setlength{\tabcolsep}{0.9pt}
	\renewcommand{\arraystretch}{1}
	\setlength{\aboverulesep}{0.25ex}
	\setlength{\belowrulesep}{0.25ex}
	
	\begin{tabular*}{\textwidth}{
			@{}ll@{\extracolsep{\fill}}cccccccccc@{}
		}
		\toprule
		Method & Publication & Clean & Fog & Rain & Snow & Frost &
		Wind & Bright. & Contr. & Night & Over-exp. \\
		\midrule
		
		FSRA
		& TCSVT'22~\cite{dai2021transformer}
		& 82.25/84.82
		& 26.13/30.03
		& 21.09/24.15
		& 20.33/24.03
		& 38.35/42.66
		& 64.08/68.32
		& 41.77/46.01
		& 40.44/44.24
		& 52.32/56.60
		& 35.66/40.10 \\
		
		MEAN
		& TGRS'25~\cite{chen2025multi}
		& 90.81/92.32
		& 81.79/86.47
		& 31.18/37.86
		& 19.67/26.46
		& 56.44/63.60
		& 72.29/78.97
		& 66.96/73.52
		& 78.41/83.32
		& 79.98/84.50
		& 65.08/71.98 \\
		
		MuSe-Net
		& PR'24~\cite{wang2024multiple}
		& 74.48/77.83
		& 46.93/51.57
		& 51.07/55.74
		& 57.74/62.16
		& 51.89/56.47
		& 57.83/62.34
		& 52.53/57.14
		& 33.15/36.65
		& 54.21/58.66
		& 49.72/54.42 \\
		
		Sample4Geo
		& ICCV'23~\cite{deuser2023sample4geo}
		& 92.65/93.81
		& 72.18/75.45
		& 41.30/44.81
		& 36.43/41.44
		& 63.77/67.51
		& 77.03/80.19
		& 70.16/73.66
		& 70.86/73.82
		& 79.26/81.75
		& 61.90/65.75 \\
		
		MCCG
		& TCSVT'24~\cite{shen2023mccg}
		& 89.64/91.32
		& 77.04/80.01
		& 39.37/43.49
		& 34.08/38.40
		& 56.13/60.15
		& 64.65/68.74
		& 71.49/74.98
		& 73.93/76.83
		& 76.30/79.10
		& 68.01/71.68 \\
		
		CAMP
		& TGRS'24~\cite{wu2024camp}
		& 94.46/95.38
		& 83.11/87.63
		& 49.74/\secondres{57.09}
		& 38.54/47.33
		& 66.90/73.72
		& 81.42/86.61
		& 73.09/79.12
		& 85.77/89.88
		& 82.79/86.96
		& 67.77/74.64 \\
		
		DAC
		& TCSVT'24~\cite{xia2024enhancing}
		& 94.67/95.50
		& \bestres{89.33}/\bestres{90.85}
		& \secondres{53.20}/56.74
		& 35.11/40.22
		& \bestres{72.02}/\bestres{75.17}
		& 84.12/86.43
		& 80.88/83.33
		& \bestres{91.53}/\bestres{92.86}
		& 87.37/88.98
		& \bestres{78.24}/\bestres{80.91} \\
		
		QDFL
		& TGRS'25~\cite{hu2025query}
		& \secondres{95.00}/\secondres{95.83}
		& 86.53/88.57
		& 51.97/55.77
		& \secondres{61.81}/\secondres{65.91}
		& 63.85/67.32
		& \secondres{90.52}/\secondres{91.99}
		& \secondres{84.80}/\secondres{87.00}
		& 89.41/90.99
		& \secondres{88.97}/\secondres{90.48}
		& 76.93/79.77 \\
		
		\textbf{Ours}
		& --
		& \bestres{95.42}/\bestres{96.36}
		& \secondres{87.51}/\secondres{89.37}
		& \bestres{61.46}/\bestres{64.97}
		& \bestres{66.14}/\bestres{70.06}
		& \secondres{71.60}/\secondres{74.73}
		& \bestres{91.45}/\bestres{92.81}
		& \bestres{85.04}/\bestres{87.11}
		& \secondres{90.98}/\secondres{92.37}
		& \bestres{90.91}/\bestres{92.19}
		& \secondres{77.28}/\secondres{79.95} \\
		
		\bottomrule
	\end{tabular*}
	
	\vspace{0.6mm}
	
	\begin{tabular*}{\textwidth}{
			@{}ll@{\extracolsep{\fill}}cccccccccc@{}
		}
		\toprule
		Method & Publication & Pixel. & JPEG & Gau. & Shot & Imp. &
		Spec. & Defocus & Glass & Motion & Zoom \\
		\midrule
		
		FSRA
		& TCSVT'22~\cite{dai2021transformer}
		& 50.26/54.26
		& 55.48/59.24
		& 41.38/44.85
		& 58.49/62.94
		& 41.15/45.66
		& 51.17/55.32
		& 34.03/38.78
		& 57.92/62.61
		& 29.19/33.57
		& 15.44/18.94 \\
		
		MEAN
		& TGRS'25~\cite{chen2025multi}
		& 43.00/49.21
		& 61.04/66.61
		& 17.12/22.10
		& 24.65/31.01
		& 14.45/20.07
		& 22.84/29.07
		& 28.07/34.01
		& 62.83/70.28
		& 14.88/21.18
		& 13.16/19.07 \\
		
		MuSe-Net
		& PR'24~\cite{wang2024multiple}
		& 51.38/56.00
		& 50.31/54.71
		& 38.10/42.36
		& 49.92/54.71
		& 54.52/59.10
		& 41.35/45.96
		& 28.04/32.37
		& 46.82/51.69
		& 32.76/37.65
		& 26.23/30.61 \\
		
		Sample4Geo
		& ICCV'23~\cite{deuser2023sample4geo}
		& 55.89/60.05
		& 65.32/68.44
		& 31.51/34.79
		& 42.90/47.18
		& 33.37/37.67
		& 36.58/40.47
		& 38.92/43.22
		& 69.93/73.76
		& 31.60/36.57
		& \bestres{46.38}/\bestres{51.07} \\
		
		MCCG
		& TCSVT'24~\cite{shen2023mccg}
		& 53.45/57.34
		& 67.06/70.37
		& 38.79/41.74
		& 53.95/58.05
		& 43.83/48.32
		& 47.76/51.55
		& 18.65/22.84
		& 43.45/48.84
		& 16.16/19.98
		& 15.32/18.45 \\
		
		CAMP
		& TGRS'24~\cite{wu2024camp}
		& 52.48/59.93
		& 65.83/71.37
		& 34.73/40.92
		& 48.45/57.01
		& 31.44/39.30
		& 41.98/49.28
		& 34.20/41.80
		& 72.05/78.89
		& 32.16/41.12
		& 31.57/38.55 \\
		
		DAC
		& TCSVT'24~\cite{xia2024enhancing}
		& 59.29/62.99
		& 72.20/74.80
		& 41.78/45.66
		& 55.23/59.83
		& 38.24/42.93
		& 47.40/51.85
		& 40.79/44.25
		& 76.45/79.51
		& 36.57/40.97
		& \secondres{37.18}/\secondres{41.20} \\
		
		QDFL
		& TGRS'25~\cite{hu2025query}
		& \secondres{76.71}/\secondres{79.43}
		& \secondres{78.12}/\secondres{80.61}
		& \secondres{49.99}/\secondres{53.37}
		& \secondres{66.86}/\secondres{70.53}
		& \secondres{59.34}/\secondres{63.70}
		& \secondres{54.99}/\secondres{58.82}
		& \secondres{68.35}/\secondres{71.89}
		& \secondres{87.78}/\secondres{89.64}
		& \secondres{64.44}/\secondres{68.23}
		& 32.62/35.52 \\
		
		\textbf{Ours}
		& --
		& \bestres{79.33}/\bestres{81.84}
		& \bestres{82.16}/\bestres{84.38}
		& \bestres{53.06}/\bestres{56.41}
		& \bestres{71.42}/\bestres{74.92}
		& \bestres{67.19}/\bestres{71.14}
		& \bestres{61.43}/\bestres{65.16}
		& \bestres{73.96}/\bestres{77.17}
		& \bestres{88.97}/\bestres{90.66}
		& \bestres{68.65}/\bestres{72.24}
		& 36.72/40.00 \\
		
		\bottomrule
	\end{tabular*}
	
	\vspace{0.6mm}
	
	\begin{tabular*}{\textwidth}{
			@{}ll@{\extracolsep{\fill}}ccccccccc@{}
		}
		\toprule
		Method & Publication & Dark+Noise & Fog+Pix. & Rain+Motion &
		Fog+Rain & Fog+Snow & Rain+Snow & Fog+Rain+Snow &
		Dark+Rain+Fog & All-27 \\
		\midrule
		
		FSRA
		& TCSVT'22~\cite{dai2021transformer}
		& 42.47/45.90
		& 25.29/28.84
		& 24.29/28.35
		& 16.89/20.44
		& 16.77/19.87
		& 26.89/31.27
		& 4.59/6.49
		& 7.53/9.58
		& 34.79/38.63 \\
		
		MEAN
		& TGRS'25~\cite{chen2025multi}
		& 23.68/30.69
		& 34.78/40.02
		& 11.36/16.44
		& 52.55/59.89
		& 51.59/60.03
		& 37.83/46.44
		& 9.89/15.45
		& 22.30/28.28
		& 40.66/46.91 \\
		
		MuSe-Net
		& PR'24~\cite{wang2024multiple}
		& 38.51/42.48
		& 42.48/47.15
		& 31.97/36.89
		& 48.89/53.71
		& 44.28/49.03
		& 59.19/63.61
		& \bestres{37.69}/\bestres{42.55}
		& 36.18/41.00
		& 44.95/49.51 \\
		
		Sample4Geo
		& ICCV'23~\cite{deuser2023sample4geo}
		& 38.25/41.87
		& 39.69/43.03
		& 28.25/33.09
		& 54.50/58.68
		& 47.29/51.90
		& 48.40/53.33
		& 12.72/16.19
		& 33.93/37.81
		& 49.20/53.09 \\
		
		MCCG
		& TCSVT'24~\cite{shen2023mccg}
		& 46.88/50.07
		& 40.81/44.66
		& 15.62/19.18
		& 61.96/65.90
		& 52.69/57.04
		& 41.55/46.16
		& 18.05/21.58
		& 32.60/36.30
		& 47.02/50.81 \\
		
		CAMP
		& TGRS'24~\cite{wu2024camp}
		& 39.49/46.20
		& 40.81/46.78
		& 27.15/35.38
		& 69.48/76.04
		& 59.11/67.20
		& 54.58/63.38
		& 20.75/28.35
		& 42.15/\secondres{49.63}
		& 52.87/59.78 \\
		
		DAC
		& TCSVT'24~\cite{xia2024enhancing}
		& 45.37/49.29
		& 43.23/46.41
		& 30.40/34.68
		& \secondres{74.07}/\secondres{77.04}
		& \secondres{64.51}/\secondres{68.34}
		& 57.68/61.86
		& 23.86/27.94
		& \bestres{48.46}/\bestres{52.38}
		& 57.95/61.39 \\
		
		QDFL
		& TGRS'25~\cite{hu2025query}
		& \secondres{58.42}/\secondres{61.66}
		& \secondres{61.33}/\secondres{64.49}
		& \secondres{52.35}/\secondres{56.16}
		& 69.45/72.84
		& 63.98/67.82
		& \secondres{60.94}/\secondres{64.94}
		& 27.83/32.05
		& 37.34/41.35
		& \secondres{65.39}/\secondres{68.55} \\
		
		\textbf{Ours}
		& --
		& \bestres{63.17}/\bestres{66.34}
		& \bestres{63.69}/\bestres{66.75}
		& \bestres{56.38}/\bestres{60.23}
		& \bestres{75.14}/\bestres{78.11}
		& \bestres{69.58}/\bestres{73.08}
		& \bestres{70.45}/\bestres{73.96}
		& \secondres{35.81}/\secondres{40.16}
		& \secondres{43.65}/47.64
		& \bestres{69.75}/\bestres{72.73} \\
		
		\bottomrule
	\end{tabular*}
	
	\vspace{-0.5mm}
\end{table*}

\begin{table*}[tb]
	\centering
	\caption{
		Comparison on University-1652-Deg for Satellite $\rightarrow$ Drone retrieval.
		Entries are R@1/AP averaged over severity levels.
		All-27 denotes the average over all 27 corrupted types and excludes Clean.
		Best and second-best results are marked in
		\textcolor{red}{red} and \textcolor{blue}{blue}, respectively.
	}
	\label{tab:u1652_sat_to_drone_all}
	
	\renewcommand{\baselinestretch}{1}
	\fontsize{7pt}{8pt}\selectfont
	\setlength{\tabcolsep}{0.9pt}
	\renewcommand{\arraystretch}{1}
	\setlength{\aboverulesep}{0.25ex}
	\setlength{\belowrulesep}{0.25ex}
	
	\begin{tabular*}{\textwidth}{
			@{}ll@{\extracolsep{\fill}}cccccccccc@{}
		}
		\toprule
		Method & Publication & Clean & Fog & Rain & Snow & Frost &
		Wind & Bright. & Contr. & Night & Over-exp. \\
		\midrule
		
		FSRA
		& TCSVT'22~\cite{dai2021transformer}
		& 87.87/81.53
		& 64.62/36.95
		& 61.58/27.01
		& 62.96/28.11
		& 71.56/47.00
		& 82.17/66.83
		& 56.68/48.57
		& 60.82/49.81
		& 83.12/56.13
		& 61.48/43.28 \\
		
		MEAN
		& TGRS'25~\cite{chen2025multi}
		& 96.01/92.08
		& 93.53/81.32
		& 74.89/36.49
		& 65.67/26.17
		& 81.65/61.65
		& 90.73/70.56
		& 82.41/71.50
		& 88.73/77.54
		& 94.06/78.78
		& 84.50/67.24 \\
		
		MuSe-Net
		& PR'24~\cite{wang2024multiple}
		& 88.02/75.10
		& 80.31/60.76
		& 79.89/62.21
		& 78.32/59.77
		& 76.08/57.21
		& 80.69/62.22
		& 75.70/57.82
		& 81.12/61.76
		& 80.79/60.91
		& 75.18/54.22 \\
		
		Sample4Geo
		& ICCV'23~\cite{deuser2023sample4geo}
		& 95.14/91.39
		& 91.30/72.81
		& 80.08/43.85
		& 73.80/39.71
		& 85.64/67.60
		& 91.44/77.55
		& 82.55/72.17
		& 84.69/71.51
		& 93.91/78.30
		& 82.60/63.56 \\
		
		MCCG
		& TCSVT'24~\cite{shen2023mccg}
		& 94.30/89.39
		& 91.73/77.02
		& 81.08/50.75
		& 77.65/45.23
		& 82.88/64.86
		& 87.97/69.85
		& 85.31/74.29
		& 87.30/75.82
		& 92.96/75.05
		& 86.64/69.92 \\
		
		CAMP
		& TGRS'24~\cite{wu2024camp}
		& 96.15/92.72
		& 94.44/82.76
		& 85.64/51.50
		& 78.65/41.81
		& 88.64/71.15
		& 93.15/80.13
		& 88.06/76.77
		& 93.96/85.52
		& 95.05/81.50
		& 87.35/70.02 \\
		
		DAC
		& TCSVT'24~\cite{xia2024enhancing}
		& 96.43/93.79
		& \bestres{95.86}/\bestres{87.85}
		& 87.87/53.95
		& 79.41/43.21
		& 90.30/74.47
		& 95.05/84.18
		& 91.06/83.21
		& \bestres{95.82}/\bestres{90.93}
		& \bestres{96.10}/85.96
		& \bestres{91.87}/\bestres{79.89} \\
		
		QDFL
		& TGRS'25~\cite{hu2025query}
		& \bestres{97.15}/\secondres{94.57}
		& \secondres{95.05}/\secondres{86.60}
		& \secondres{89.44}/\secondres{63.11}
		& \secondres{90.82}/\secondres{70.82}
		& \secondres{90.54}/\secondres{76.70}
		& \secondres{95.20}/\secondres{89.58}
		& \bestres{93.06}/\secondres{87.59}
		& 95.01/89.74
		& 95.39/\secondres{87.62}
		& \secondres{91.58}/\secondres{79.73} \\
		
		\textbf{Ours}
		& --
		& \secondres{97.10}/\bestres{95.01}
		& 94.48/86.40
		& \bestres{91.06}/\bestres{68.75}
		& \bestres{92.44}/\bestres{73.81}
		& \bestres{91.49}/\bestres{79.27}
		& \bestres{95.48}/\bestres{90.09}
		& \secondres{92.58}/\bestres{87.67}
		& \secondres{95.44}/\secondres{90.86}
		& \secondres{96.01}/\bestres{88.70}
		& 91.44/\bestres{79.89} \\
		
		\bottomrule
	\end{tabular*}
	
	\vspace{0.6mm}
	
	\begin{tabular*}{\textwidth}{
			@{}ll@{\extracolsep{\fill}}cccccccccc@{}
		}
		\toprule
		Method & Publication & Pixel. & JPEG & Gau. & Shot & Imp. &
		Spec. & Defocus & Glass & Motion & Zoom \\
		\midrule
		
		FSRA
		& TCSVT'22~\cite{dai2021transformer}
		& 75.23/58.93
		& 73.61/61.05
		& 62.29/51.10
		& 77.27/66.45
		& 68.90/57.62
		& 71.61/59.53
		& 55.16/39.15
		& 77.03/62.17
		& 56.87/35.23
		& 38.76/31.41 \\
		
		MEAN
		& TGRS'25~\cite{chen2025multi}
		& 71.09/46.44
		& 78.27/64.22
		& 36.28/28.18
		& 50.83/39.22
		& 41.32/30.33
		& 46.46/35.44
		& 42.13/29.64
		& 80.36/63.17
		& 41.65/16.59
		& 31.62/21.34 \\
		
		MuSe-Net
		& PR'24~\cite{wang2024multiple}
		& 76.42/55.75
		& 68.19/51.11
		& 68.24/50.17
		& 75.23/58.08
		& 76.56/60.47
		& 69.62/52.19
		& 62.20/41.54
		& 76.89/57.20
		& 67.24/43.36
		& 52.78/37.82 \\
		
		Sample4Geo
		& ICCV'23~\cite{deuser2023sample4geo}
		& 81.55/60.41
		& 82.74/69.09
		& 51.16/39.01
		& 68.66/52.51
		& 63.10/47.69
		& 60.58/45.86
		& 59.72/42.45
		& 86.21/71.62
		& 63.58/37.84
		& \bestres{67.67}/\bestres{53.96} \\
		
		MCCG
		& TCSVT'24~\cite{shen2023mccg}
		& 80.65/61.47
		& 82.74/68.45
		& 60.63/49.31
		& 80.17/66.18
		& 76.94/62.74
		& 72.33/58.07
		& 47.93/33.20
		& 77.56/61.19
		& 53.21/29.01
		& 43.84/34.49 \\
		
		CAMP
		& TGRS'24~\cite{wu2024camp}
		& 79.98/56.86
		& 83.40/69.65
		& 61.06/46.98
		& 80.88/63.41
		& 64.10/48.26
		& 71.61/54.78
		& 60.72/41.53
		& 88.45/73.17
		& 68.85/39.62
		& 59.72/45.84 \\
		
		DAC
		& TCSVT'24~\cite{xia2024enhancing}
		& 84.36/63.00
		& 88.87/75.67
		& 70.09/53.92
		& 85.92/70.30
		& 74.94/58.88
		& 79.36/62.43
		& 64.34/44.12
		& 92.11/78.38
		& 69.04/40.59
		& 62.96/48.50 \\
		
		QDFL
		& TGRS'25~\cite{hu2025query}
		& \secondres{91.25}/\secondres{78.47}
		& \secondres{90.54}/\secondres{81.11}
		& \secondres{79.84}/\secondres{68.00}
		& \secondres{90.39}/\secondres{82.70}
		& \secondres{91.30}/\secondres{81.66}
		& \secondres{84.69}/\secondres{75.06}
		& \secondres{84.78}/\secondres{70.31}
		& \secondres{93.96}/\secondres{88.46}
		& \secondres{88.97}/\secondres{67.22}
		& 64.86/49.57 \\
		
		\textbf{Ours}
		& --
		& \bestres{92.34}/\bestres{80.95}
		& \bestres{92.34}/\bestres{83.97}
		& \bestres{80.27}/\bestres{68.55}
		& \bestres{91.96}/\bestres{84.15}
		& \bestres{92.87}/\bestres{84.12}
		& \bestres{88.40}/\bestres{77.77}
		& \bestres{87.59}/\bestres{72.86}
		& \bestres{94.44}/\bestres{88.96}
		& \bestres{89.54}/\bestres{68.66}
		& \secondres{66.57}/\secondres{51.97} \\
		
		\bottomrule
	\end{tabular*}
	
	\vspace{0.6mm}
	
	\begin{tabular*}{\textwidth}{
			@{}ll@{\extracolsep{\fill}}ccccccccc@{}
		}
		\toprule
		Method & Publication & Dark+Noise & Fog+Pix. & Rain+Motion &
		Fog+Rain & Fog+Snow & Rain+Snow & Fog+Rain+Snow &
		Dark+Rain+Fog & All-27 \\
		\midrule
		
		FSRA
		& TCSVT'22~\cite{dai2021transformer}
		& 64.19/53.47
		& 54.11/37.94
		& 53.02/34.22
		& 60.77/26.62
		& 50.88/25.40
		& 68.71/36.58
		& 35.57/10.19
		& 49.93/14.07
		& 62.92/43.14 \\
		
		MEAN
		& TGRS'25~\cite{chen2025multi}
		& 48.17/37.39
		& 55.59/37.54
		& 34.33/15.17
		& 90.68/52.54
		& 85.40/55.70
		& 80.79/43.79
		& 57.35/16.38
		& 82.93/25.76
		& 67.09/45.56 \\
		
		MuSe-Net
		& PR'24~\cite{wang2024multiple}
		& 68.52/50.65
		& 74.13/53.25
		& 67.28/42.96
		& 80.93/63.41
		& 78.75/59.17
		& 79.50/62.80
		& 76.99/\bestres{55.26}
		& 78.32/\bestres{54.72}
		& 74.29/55.07 \\
		
		Sample4Geo
		& ICCV'23~\cite{deuser2023sample4geo}
		& 59.01/46.97
		& 63.72/43.94
		& 59.15/35.19
		& 88.92/53.80
		& 78.89/48.82
		& 82.45/52.44
		& 53.21/17.08
		& 84.45/33.29
		& 74.84/53.30 \\
		
		MCCG
		& TCSVT'24~\cite{shen2023mccg}
		& 68.47/55.34
		& 69.33/49.84
		& 53.40/29.29
		& 90.73/63.73
		& 85.40/59.93
		& 82.74/54.62
		& 66.91/28.85
		& 85.31/38.39
		& 75.99/55.81 \\
		
		CAMP
		& TGRS'24~\cite{wu2024camp}
		& 67.62/52.53
		& 64.10/44.45
		& 60.10/33.62
		& 92.72/66.41
		& 87.40/60.94
		& 86.45/59.33
		& 66.95/26.06
		& \secondres{90.25}/42.02
		& 79.23/58.02 \\
		
		DAC
		& TCSVT'24~\cite{xia2024enhancing}
		& 74.37/59.18
		& 67.62/45.90
		& 62.34/35.72
		& \bestres{94.53}/70.85
		& 89.73/65.34
		& 88.30/59.88
		& 68.62/27.81
		& \bestres{92.39}/46.97
		& 82.71/62.63 \\
		
		QDFL
		& TGRS'25~\cite{hu2025query}
		& \secondres{84.07}/\secondres{72.92}
		& \secondres{82.60}/\secondres{65.88}
		& \secondres{82.22}/\secondres{57.66}
		& 93.87/\secondres{71.81}
		& \secondres{91.77}/\secondres{70.71}
		& \secondres{91.68}/\secondres{71.05}
		& \secondres{80.93}/42.79
		& 88.78/44.54
		& \secondres{88.61}/\secondres{73.01} \\
		
		\textbf{Ours}
		& --
		& \bestres{86.40}/\bestres{75.58}
		& \bestres{83.98}/\bestres{68.03}
		& \bestres{83.83}/\bestres{59.25}
		& \secondres{94.34}/\bestres{74.49}
		& \bestres{92.68}/\bestres{73.72}
		& \bestres{92.87}/\bestres{75.90}
		& \bestres{84.02}/\secondres{48.00}
		& 90.01/\secondres{48.08}
		& \bestres{89.81}/\bestres{75.20} \\
		
		\bottomrule
	\end{tabular*}
	
	\vspace{-0.5mm}
\end{table*}

\begin{table*}[tb]
	\centering
	\caption{
		Comparison on SUES-200-Deg for Drone $\rightarrow$ Satellite retrieval.
		Entries are R@1/AP averaged over severity levels and four UAV heights.
		All-27 denotes the average over all 27 corrupted types and excludes Clean.
		Best and second-best results are marked in
		\textcolor{red}{red} and \textcolor{blue}{blue}, respectively.
	}
	\label{tab:sues_drone_to_sat_all}
	
	\renewcommand{\baselinestretch}{1}
	\fontsize{7pt}{8pt}\selectfont
	\setlength{\tabcolsep}{0.9pt}
	\renewcommand{\arraystretch}{1}
	
	\setlength{\heavyrulewidth}{0.4pt}
	\setlength{\lightrulewidth}{0.4pt}
	\setlength{\cmidrulewidth}{0.4pt}
	\setlength{\aboverulesep}{0.25ex}
	\setlength{\belowrulesep}{0.25ex}
	
	\begin{tabular*}{\textwidth}{
			@{}ll@{\extracolsep{\fill}}cccccccccc@{}
		}
		\toprule
		Method & Publication & Clean & Fog & Rain & Snow & Frost &
		Wind & Bright. & Contr. & Night & Over-exp. \\
		\midrule
		
		FSRA
		& TCSVT'22~\cite{dai2021transformer}
		& 83.47/85.92
		& 33.89/45.43
		& 26.28/36.77
		& 19.59/29.82
		& 43.39/55.54
		& 68.98/78.94
		& 50.10/59.72
		& 48.35/56.65
		& 58.43/67.91
		& 49.70/61.97 \\
		
		MEAN
		& TGRS'25~\cite{chen2025multi}
		& \secondres{98.10}/\secondres{98.50}
		& 86.01/88.17
		& 37.67/43.38
		& 28.87/34.53
		& 59.79/64.44
		& 81.22/84.03
		& 79.74/82.45
		& 79.31/81.85
		& 89.42/91.08
		& 81.60/84.10 \\
		
		CCR
		& TCSVT'24~\cite{du2024ccr}
		& 93.22/94.53
		& 87.57/89.89
		& 52.46/57.59
		& 48.16/53.36
		& 70.44/74.59
		& 76.72/80.27
		& 76.70/79.75
		& 86.89/89.20
		& 86.91/89.08
		& 80.58/83.50 \\
		
		MCCG
		& TCSVT'24~\cite{shen2023mccg}
		& 90.12/92.03
		& 81.41/84.61
		& 48.04/53.80
		& 39.95/46.00
		& 63.08/67.70
		& 81.45/84.76
		& 73.57/77.47
		& 80.39/83.53
		& 83.63/86.35
		& 76.81/80.43 \\
		
		DAC
		& TCSVT'24~\cite{xia2024enhancing}
		& 97.52/98.07
		& 86.12/88.41
		& 51.29/56.00
		& 47.39/52.70
		& 70.86/74.73
		& 84.63/87.22
		& 83.49/86.01
		& 89.29/91.32
		& 91.89/93.41
		& 81.64/84.35 \\
		
		CAMP
		& TGRS'24~\cite{wu2024camp}
		& 97.60/98.11
		& \secondres{90.81}/92.41
		& 60.98/65.14
		& 52.38/57.65
		& \secondres{80.52}/83.25
		& 89.86/91.66
		& 84.43/86.65
		& \bestres{94.19}/95.31
		& \secondres{94.90}/95.82
		& 84.68/86.88 \\
		
		Sample4Geo
		& ICCV'23~\cite{deuser2023sample4geo}
		& 96.86/97.45
		& 86.50/88.63
		& \bestres{64.66}/\secondres{68.69}
		& 62.14/66.93
		& \bestres{83.65}/\bestres{86.16}
		& 90.34/92.04
		& 82.12/84.66
		& 87.50/89.50
		& 93.03/94.18
		& 81.92/84.48 \\
		
		QDFL
		& TGRS'25~\cite{hu2025query}
		& 97.71/98.29
		& 90.37/\secondres{94.23}
		& 56.87/66.89
		& \secondres{63.02}/\secondres{73.17}
		& 72.00/79.95
		& \secondres{93.61}/\secondres{96.23}
		& \secondres{88.84}/\secondres{92.98}
		& \secondres{92.27}/\bestres{95.36}
		& 94.48/\secondres{96.68}
		& \secondres{87.39}/\secondres{91.80} \\
		
		\textbf{Ours}
		& --
		& \bestres{98.24}/\bestres{98.67}
		& \bestres{92.33}/\bestres{95.47}
		& \secondres{61.01}/\bestres{70.47}
		& \bestres{70.34}/\bestres{78.92}
		& 76.49/\secondres{83.60}
		& \bestres{94.73}/\bestres{96.98}
		& \bestres{90.68}/\bestres{94.38}
		& 92.25/\secondres{95.32}
		& \bestres{95.42}/\bestres{97.37}
		& \bestres{89.01}/\bestres{92.99} \\
	\end{tabular*}
	
	\vspace{0.6mm}
	
	\begin{tabular*}{\textwidth}{
			@{}ll@{\extracolsep{\fill}}cccccccccc@{}
		}
		\toprule
		Method & Publication & Pixel. & JPEG & Gau. & Shot & Imp. &
		Spec. & Defocus & Glass & Motion & Zoom \\
		\midrule
		
		FSRA
		& TCSVT'22~\cite{dai2021transformer}
		& 52.35/63.01
		& 64.09/74.34
		& 50.71/61.33
		& 64.54/75.27
		& 47.25/60.24
		& 57.03/68.44
		& 30.88/42.20
		& 58.80/70.25
		& 29.42/42.04
		& 19.79/30.09 \\
		
		MEAN
		& TGRS'25~\cite{chen2025multi}
		& 47.90/52.45
		& 79.92/82.81
		& 33.61/38.41
		& 46.35/51.52
		& 40.77/46.06
		& 43.31/48.15
		& 40.20/44.60
		& 76.30/79.64
		& 25.02/30.81
		& 35.62/40.59 \\
		
		CCR
		& TCSVT'24~\cite{du2024ccr}
		& 45.95/49.76
		& 75.18/78.76
		& 31.70/35.90
		& 43.46/48.61
		& 49.51/54.40
		& 40.68/45.41
		& 35.54/39.66
		& 69.10/73.33
		& 21.70/25.91
		& 25.75/30.12 \\
		
		MCCG
		& TCSVT'24~\cite{shen2023mccg}
		& 67.74/72.30
		& 80.61/84.07
		& 53.71/58.48
		& 70.35/75.08
		& 63.78/69.18
		& 63.84/68.80
		& 41.57/47.31
		& 70.06/74.68
		& 35.37/41.41
		& 27.38/32.31 \\
		
		DAC
		& TCSVT'24~\cite{xia2024enhancing}
		& 56.42/61.10
		& 84.25/86.77
		& 49.45/53.53
		& 65.43/69.62
		& 55.80/60.60
		& 58.15/62.44
		& 45.56/50.18
		& 78.78/82.17
		& 31.76/37.20
		& 41.04/45.71 \\
		
		CAMP
		& TGRS'24~\cite{wu2024camp}
		& 61.99/66.38
		& \bestres{87.79}/\secondres{89.66}
		& 56.16/60.28
		& 73.34/77.20
		& 67.51/71.74
		& 65.60/69.66
		& 50.11/54.56
		& 85.08/87.51
		& 41.37/46.84
		& 46.31/50.99 \\
		
		Sample4Geo
		& ICCV'23~\cite{deuser2023sample4geo}
		& 62.42/66.78
		& 85.93/88.05
		& \secondres{62.70}/66.70
		& 79.15/82.34
		& \bestres{75.54}/\secondres{79.10}
		& \secondres{71.17}/74.93
		& 51.67/56.41
		& 83.61/86.33
		& 47.44/52.78
		& \bestres{50.49}/\secondres{55.13} \\
		
		QDFL
		& TGRS'25~\cite{hu2025query}
		& \secondres{78.81}/\secondres{84.87}
		& 84.81/89.51
		& 62.12/\secondres{70.10}
		& \secondres{81.18}/\secondres{87.77}
		& 69.22/78.02
		& \secondres{71.17}/\secondres{78.81}
		& \secondres{66.18}/\secondres{75.25}
		& \secondres{90.87}/\secondres{94.54}
		& \secondres{64.37}/\secondres{73.81}
		& 44.93/53.42 \\
		
		\textbf{Ours}
		& --
		& \bestres{82.89}/\bestres{88.22}
		& \secondres{87.30}/\bestres{91.57}
		& \bestres{63.91}/\bestres{71.32}
		& \bestres{83.63}/\bestres{89.25}
		& \secondres{73.20}/\bestres{81.12}
		& \bestres{72.85}/\bestres{80.20}
		& \bestres{67.21}/\bestres{75.57}
		& \bestres{91.87}/\bestres{95.28}
		& \bestres{69.80}/\bestres{78.32}
		& \secondres{47.95}/\bestres{56.50} \\
	\end{tabular*}
	
	\vspace{0.6mm}
	
	\begin{tabular*}{\textwidth}{
			@{}ll@{\extracolsep{\fill}}ccccccccc@{}
		}
		\toprule
		Method & Publication & Dark+Noise & Fog+Pix. & Rain+Motion &
		Fog+Rain & Fog+Snow & Rain+Snow & Fog+Rain+Snow &
		Dark+Rain+Fog & All-27 \\
		\midrule
		
		FSRA
		& TCSVT'22~\cite{dai2021transformer}
		& 49.13/58.99
		& 32.24/42.96
		& 26.79/39.24
		& 24.43/36.19
		& 24.67/34.83
		& 29.46/42.89
		& 8.33/15.72
		& 11.17/19.43
		& 39.99/50.75 \\
		
		MEAN
		& TGRS'25~\cite{chen2025multi}
		& 42.39/46.97
		& 42.26/46.47
		& 20.00/25.50
		& 60.24/65.08
		& 57.26/62.04
		& 40.62/46.36
		& 16.49/21.62
		& 28.88/34.11
		& 51.88/56.19 \\
		
		CCR
		& TCSVT'24~\cite{du2024ccr}
		& 44.37/49.02
		& 39.36/42.67
		& 17.40/21.54
		& 70.84/74.90
		& 68.57/73.02
		& 56.85/62.31
		& 28.53/34.07
		& 41.84/46.90
		& 54.55/58.65 \\
		
		MCCG
		& TCSVT'24~\cite{shen2023mccg}
		& 63.37/67.75
		& 54.82/59.82
		& 31.94/38.21
		& 69.89/74.31
		& 62.00/67.00
		& 48.25/54.54
		& 22.63/28.55
		& 39.40/45.18
		& 59.08/63.84 \\
		
		DAC
		& TCSVT'24~\cite{xia2024enhancing}
		& 56.68/60.54
		& 47.93/52.47
		& 24.06/29.45
		& 70.32/74.23
		& 67.17/71.35
		& 57.40/62.20
		& 29.55/34.80
		& 38.75/43.55
		& 60.93/64.89 \\
		
		CAMP
		& TGRS'24~\cite{wu2024camp}
		& 64.87/68.17
		& 55.75/59.90
		& 34.96/40.46
		& \bestres{77.26}/\secondres{80.44}
		& \secondres{74.34}/77.86
		& 68.42/72.58
		& 35.29/40.52
		& \bestres{47.39}/52.02
		& 67.64/71.17 \\
		
		Sample4Geo
		& ICCV'23~\cite{deuser2023sample4geo}
		& 68.52/71.86
		& 52.72/57.11
		& 40.72/46.27
		& 75.95/79.45
		& \bestres{74.53}/\secondres{78.10}
		& \bestres{73.76}/\secondres{77.56}
		& \bestres{40.72}/\secondres{46.12}
		& 44.93/49.95
		& 69.40/72.97 \\
		
		QDFL
		& TGRS'25~\cite{hu2025query}
		& \secondres{70.44}/\secondres{77.33}
		& \secondres{57.40}/\secondres{65.82}
		& \secondres{45.71}/\secondres{55.43}
		& 72.12/80.33
		& 68.17/77.08
		& 61.99/72.23
		& 31.56/43.37
		& 42.83/\secondres{53.80}
		& \secondres{70.47}/\secondres{77.73} \\
		
		\textbf{Ours}
		& --
		& \bestres{70.87}/\bestres{77.71}
		& \bestres{61.48}/\bestres{69.73}
		& \bestres{51.26}/\bestres{61.00}
		& \secondres{77.16}/\bestres{84.26}
		& 73.88/\bestres{81.62}
		& \secondres{70.96}/\bestres{79.44}
		& \secondres{37.67}/\bestres{49.07}
		& \secondres{45.79}/\bestres{56.50}
		& \bestres{73.78}/\bestres{80.45} \\
		
		\bottomrule
	\end{tabular*}
	
	\vspace{-0.5mm}
\end{table*}

\begin{table*}[tb]
	\centering
	\caption{
		Comparison on SUES-200-Deg for Satellite $\rightarrow$ Drone retrieval.
		Entries are R@1/AP averaged over severity levels and four UAV heights.
		All-27 denotes the average over all 27 corrupted types and excludes Clean.
		Best and second-best results are marked in
		\textcolor{red}{red} and \textcolor{blue}{blue}, respectively.
	}
	\label{tab:sues_sat_to_drone_all}
	
	\renewcommand{\baselinestretch}{1}
	\fontsize{7pt}{8pt}\selectfont
	\setlength{\tabcolsep}{0.9pt}
	\renewcommand{\arraystretch}{1}
	
	\setlength{\heavyrulewidth}{0.4pt}
	\setlength{\lightrulewidth}{0.4pt}
	\setlength{\cmidrulewidth}{0.4pt}
	\setlength{\aboverulesep}{0.25ex}
	\setlength{\belowrulesep}{0.25ex}
	
	\begin{tabular*}{\textwidth}{
			@{}ll@{\extracolsep{\fill}}cccccccccc@{}
		}
		\toprule
		Method & Publication & Clean & Fog & Rain & Snow & Frost &
		Wind & Bright. & Contr. & Night & Over-exp. \\
		\midrule
		
		FSRA
		& TCSVT'22~\cite{dai2021transformer}
		& 90.63/86.10
		& 63.75/41.88
		& 64.27/32.79
		& 61.88/33.04
		& 75.62/52.43
		& 88.54/73.29
		& 65.52/57.53
		& 63.33/55.15
		& 84.48/61.99
		& 75.62/56.19 \\
		
		MEAN
		& TGRS'25~\cite{chen2025multi}
		& \bestres{99.38}/97.33
		& 95.21/86.33
		& 75.10/40.03
		& 73.33/40.74
		& 84.38/68.11
		& 94.90/82.06
		& 88.65/83.25
		& 85.52/80.41
		& 95.94/88.76
		& 92.19/81.20 \\
		
		CCR
		& TCSVT'24~\cite{du2024ccr}
		& 96.25/94.59
		& 95.73/89.33
		& 85.21/58.10
		& 86.25/65.08
		& 87.92/76.00
		& 93.12/80.26
		& 86.56/81.44
		& 94.38/89.66
		& 95.42/88.05
		& 92.50/82.62 \\
		
		MCCG
		& TCSVT'24~\cite{shen2023mccg}
		& 95.63/93.68
		& 94.06/83.63
		& 84.06/56.07
		& 78.44/49.88
		& 86.98/69.48
		& 94.27/85.05
		& 86.46/80.00
		& 91.46/83.78
		& 96.35/85.21
		& 91.15/79.42 \\
		
		DAC
		& TCSVT'24~\cite{xia2024enhancing}
		& \secondres{98.44}/96.67
		& 94.17/85.12
		& 78.33/45.72
		& 81.67/53.73
		& 85.31/72.98
		& 93.33/81.99
		& 88.85/83.43
		& 92.08/88.23
		& 95.94/90.50
		& 90.42/80.10 \\
		
		CAMP
		& TGRS'24~\cite{wu2024camp}
		& 98.13/96.85
		& 97.19/90.37
		& 88.02/57.27
		& 89.38/60.85
		& 93.75/81.98
		& 97.08/89.36
		& 90.00/85.30
		& 97.08/93.55
		& 98.02/94.14
		& 93.85/83.20 \\
		
		Sample4Geo
		& ICCV'23~\cite{deuser2023sample4geo}
		& \secondres{98.44}/95.67
		& 95.21/85.98
		& 89.48/61.26
		& 90.52/69.26
		& 93.54/\bestres{83.13}
		& 95.94/89.32
		& 88.23/82.56
		& 92.60/86.49
		& 97.08/91.14
		& 92.29/81.31 \\
		
		QDFL
		& TGRS'25~\cite{hu2025query}
		& \bestres{99.38}/\secondres{97.79}
		& \secondres{97.71}/\secondres{92.53}
		& \secondres{91.35}/\bestres{67.88}
		& \secondres{95.21}/\secondres{79.04}
		& \secondres{94.06}/80.29
		& \secondres{97.71}/\secondres{94.59}
		& \secondres{95.42}/\secondres{92.82}
		& \bestres{97.50}/\bestres{94.11}
		& \bestres{98.96}/\secondres{95.30}
		& \secondres{95.52}/\secondres{89.81} \\
		
		\textbf{Ours}
		& --
		& \bestres{99.38}/\bestres{98.32}
		& \bestres{98.33}/\bestres{93.60}
		& \bestres{92.60}/\secondres{67.63}
		& \bestres{96.35}/\bestres{81.87}
		& \bestres{95.10}/\secondres{82.49}
		& \bestres{98.65}/\bestres{95.73}
		& \bestres{96.88}/\bestres{93.50}
		& \secondres{97.29}/\secondres{93.58}
		& \secondres{98.85}/\bestres{96.11}
		& \bestres{97.29}/\bestres{91.13} \\
	\end{tabular*}
	
	\vspace{0.6mm}
	
	\begin{tabular*}{\textwidth}{
			@{}ll@{\extracolsep{\fill}}cccccccccc@{}
		}
		\toprule
		Method & Publication & Pixel. & JPEG & Gau. & Shot & Imp. &
		Spec. & Defocus & Glass & Motion & Zoom \\
		\midrule
		
		FSRA
		& TCSVT'22~\cite{dai2021transformer}
		& 73.75/62.13
		& 81.67/70.98
		& 69.69/58.43
		& 82.19/71.19
		& 74.17/62.96
		& 77.50/66.31
		& 49.27/40.19
		& 76.04/66.21
		& 52.92/36.44
		& 39.38/31.20 \\
		
		MEAN
		& TGRS'25~\cite{chen2025multi}
		& 71.04/58.53
		& 90.62/83.91
		& 46.04/41.04
		& 61.67/54.91
		& 58.75/51.59
		& 58.65/51.60
		& 56.25/48.09
		& 88.23/81.40
		& 55.21/33.15
		& 49.79/41.37 \\
		
		CCR
		& TCSVT'24~\cite{du2024ccr}
		& 65.73/56.34
		& 87.19/80.23
		& 50.42/45.08
		& 67.50/60.37
		& 73.33/66.62
		& 63.54/56.03
		& 53.96/46.29
		& 83.85/76.86
		& 48.23/32.23
		& 51.35/41.86 \\
		
		MCCG
		& TCSVT'24~\cite{shen2023mccg}
		& 85.52/75.05
		& 92.71/83.64
		& 72.50/63.04
		& 88.12/78.59
		& 84.06/74.88
		& 81.56/72.75
		& 66.67/56.63
		& 89.06/80.20
		& 68.54/47.73
		& 51.25/42.67 \\
		
		DAC
		& TCSVT'24~\cite{xia2024enhancing}
		& 73.23/60.28
		& 88.75/83.40
		& 57.71/52.41
		& 73.33/66.50
		& 67.40/60.43
		& 66.35/61.00
		& 56.88/49.36
		& 85.73/78.37
		& 53.44/36.97
		& 52.92/45.19 \\
		
		CAMP
		& TGRS'24~\cite{wu2024camp}
		& 83.96/68.83
		& \bestres{95.21}/\secondres{88.44}
		& 68.65/60.52
		& 83.12/75.54
		& 79.90/71.60
		& 76.67/69.04
		& 63.33/54.77
		& 91.98/85.75
		& 67.29/47.19
		& 61.98/51.56 \\
		
		Sample4Geo
		& ICCV'23~\cite{deuser2023sample4geo}
		& 82.60/68.51
		& 92.81/86.14
		& 74.69/66.87
		& 88.75/81.18
		& 85.73/77.79
		& 81.25/73.93
		& 66.15/56.94
		& 91.88/85.21
		& 74.58/55.60
		& \secondres{63.54}/54.09 \\
		
		QDFL
		& TGRS'25~\cite{hu2025query}
		& \secondres{89.58}/\secondres{83.37}
		& 92.92/88.16
		& \secondres{78.23}/\bestres{70.97}
		& \bestres{93.75}/\secondres{87.38}
		& \secondres{88.85}/\secondres{82.15}
		& \secondres{86.46}/\secondres{79.81}
		& \secondres{80.62}/\bestres{72.46}
		& \secondres{95.62}/\secondres{92.76}
		& \secondres{84.48}/\secondres{68.32}
		& 63.02/\secondres{54.54} \\
		
		\textbf{Ours}
		& --
		& \bestres{93.02}/\bestres{86.24}
		& \secondres{94.27}/\bestres{89.66}
		& \bestres{78.75}/\secondres{70.85}
		& \secondres{93.54}/\bestres{87.82}
		& \bestres{89.79}/\bestres{83.86}
		& \bestres{87.40}/\bestres{79.98}
		& \bestres{81.35}/\secondres{72.17}
		& \bestres{97.08}/\bestres{93.95}
		& \bestres{89.90}/\bestres{72.73}
		& \bestres{66.46}/\bestres{56.84} \\
	\end{tabular*}
	
	\vspace{0.6mm}
	
	\begin{tabular*}{\textwidth}{
			@{}ll@{\extracolsep{\fill}}ccccccccc@{}
		}
		\toprule
		Method & Publication & Dark+Noise & Fog+Pix. & Rain+Motion &
		Fog+Rain & Fog+Snow & Rain+Snow & Fog+Rain+Snow &
		Dark+Rain+Fog & All-27 \\
		\midrule
		
		FSRA
		& TCSVT'22~\cite{dai2021transformer}
		& 67.60/57.65
		& 55.00/40.08
		& 51.98/35.61
		& 62.19/31.44
		& 51.04/29.90
		& 70.52/41.22
		& 36.88/13.46
		& 47.92/17.60
		& 65.29/48.05 \\
		
		MEAN
		& TGRS'25~\cite{chen2025multi}
		& 55.73/50.03
		& 63.65/51.03
		& 46.15/28.45
		& 92.81/57.72
		& 86.67/62.96
		& 79.27/45.83
		& 56.46/21.13
		& 80.94/28.57
		& 73.45/57.12 \\
		
		CCR
		& TCSVT'24~\cite{du2024ccr}
		& 65.10/57.76
		& 56.15/46.82
		& 43.33/27.31
		& 93.75/71.93
		& 89.69/73.97
		& 88.44/67.39
		& 71.77/39.55
		& 88.65/45.66
		& 76.26/63.07 \\
		
		MCCG
		& TCSVT'24~\cite{shen2023mccg}
		& 80.10/71.54
		& 75.62/62.07
		& 65.62/44.44
		& 93.23/71.69
		& 87.29/64.73
		& 84.06/58.29
		& 64.48/31.93
		& 87.08/43.37
		& 82.25/66.51 \\
		
		DAC
		& TCSVT'24~\cite{xia2024enhancing}
		& 66.56/59.86
		& 65.21/51.56
		& 45.10/29.43
		& 92.29/62.78
		& 86.04/66.37
		& 81.56/54.91
		& 63.44/28.81
		& 81.98/33.53
		& 76.22/61.59 \\
		
		CAMP
		& TGRS'24~\cite{wu2024camp}
		& 73.96/67.55
		& \secondres{77.29}/60.80
		& 61.77/40.50
		& \secondres{95.73}/71.41
		& \secondres{91.77}/73.13
		& 91.56/67.30
		& 73.02/35.05
		& \secondres{91.56}/42.91
		& 84.19/69.18 \\
		
		Sample4Geo
		& ICCV'23~\cite{deuser2023sample4geo}
		& 79.37/72.13
		& 73.65/58.96
		& \secondres{68.23}/49.59
		& 94.58/71.75
		& 91.25/73.87
		& \secondres{93.33}/73.05
		& 75.42/41.35
		& 87.60/42.07
		& 85.20/71.09 \\
		
		QDFL
		& TGRS'25~\cite{hu2025query}
		& \secondres{84.79}/\bestres{78.75}
		& 76.88/\secondres{67.18}
		& \secondres{68.23}/\secondres{52.28}
		& \secondres{95.73}/\secondres{77.38}
		& 91.25/\secondres{77.08}
		& 92.71/\secondres{75.06}
		& \secondres{78.65}/\secondres{45.76}
		& 88.75/\bestres{49.68}
		& \secondres{88.67}/\secondres{77.39} \\
		
		\textbf{Ours}
		& --
		& \bestres{85.83}/\secondres{78.55}
		& \bestres{79.79}/\bestres{69.62}
		& \bestres{73.85}/\bestres{57.28}
		& \bestres{96.98}/\bestres{78.83}
		& \bestres{93.54}/\bestres{78.63}
		& \bestres{95.62}/\bestres{77.48}
		& \bestres{79.27}/\bestres{45.80}
		& \bestres{91.88}/\secondres{48.38}
		& \bestres{90.36}/\bestres{78.68} \\
		
		\bottomrule
	\end{tabular*}
	
	\vspace{-0.5mm}
\end{table*}


\subsection{Datasets and Evaluation Metrics}
\label{sec:experiments_datasets_metrics}

We evaluate ReLATE on University-1652-Deg and SUES-200-Deg under
the clean-training corrupted-testing protocol defined in Section~\ref{sec:benchmark}.
D2S uses degraded UAV queries and a clean satellite gallery, whereas
S2D uses clean satellite queries and a degraded UAV gallery.
All 27 corruption types are evaluated at three severity levels; unless
otherwise specified, SUES-200-Deg results are additionally
macro-averaged over H150, H200, H250, and H300.

We report standard retrieval metrics, including Recall at rank $K$ (R@$K$) and mean Average Precision (mAP). Given $N_q$ query images, R@$K$ is computed as
\begin{equation}
	\mathrm{R@}K =
    \frac{1}{N_q}
    \sum_{i=1}^{N_q}
	\mathbb{I}
	\left[
	\operatorname{rank}_{i}^{+} \leq K
	\right],
\end{equation}
where $\operatorname{rank}_{i}^{+}$ denotes the rank of the highest-ranked correct gallery image for the $i$-th query, and $\mathbb{I}[\cdot]$ is the indicator function. We mainly use R@1 in the main comparison tables because it directly reflects the top-match localization accuracy.

For average precision, let $\mathcal{G}_{i}^{+}$ denote the set of correct gallery images for query $i$, and let $\rho_i(k)\in\{0,1\}$ indicate whether the $k$-th retrieved gallery image is correct. The average precision of query $i$ is defined as
\begin{equation}
	\mathrm{AP}_{i} =
	\frac{1}{|\mathcal{G}_{i}^{+}|}
	\sum_{k=1}^{|\mathcal{G}|}
	P_i(k)\,\rho_i(k),
\end{equation}
where $P_i(k)$ is the precision among the top-$k$ retrieved results. The mean Average Precision is then computed as
\begin{equation}
	\mathrm{mAP} =
    \frac{1}{N_q}
    \sum_{i=1}^{N_q}
	\mathrm{AP}_{i}.
\end{equation}
For compactness, we denote mAP as AP in the result tables.

To avoid bias from different corruption groups or UAV heights, all reported corrupted averages are computed as macro-averages over fixed evaluation subsets. For a corruption type $c$, the per-corruption score is averaged over the three severity levels:
\begin{equation}
	M(c) =
	\frac{1}{|\mathcal{S}|}
	\sum_{s \in \mathcal{S}}
	M(c,s),
	\quad
	\mathcal{S}=\{1,2,3\},
\end{equation}
where $M$ denotes a retrieval metric such as R@1 or AP. For SUES-200-Deg, the score is further averaged over the four UAV heights:
\begin{equation}
	\begin{aligned}
		M_{\mathrm{SUES}}(c)
		&=
		\frac{1}{|\mathcal{H}||\mathcal{S}|}
		\sum_{h \in \mathcal{H}}
		\sum_{s \in \mathcal{S}}
		M(h,c,s), \\
		\mathcal{H}
		&=
		\{\mathrm{H150}, \mathrm{H200}, \mathrm{H250}, \mathrm{H300}\}.
	\end{aligned}
\end{equation}
The All-27 column in each main comparison table denotes the average over the corrupted types included in that table and does not include the clean subset. Clean performance is reported separately to show whether robustness gains are achieved without sacrificing standard clean-image retrieval accuracy.

\subsection{Experimental Setup}
\label{sec:experimental_setup}

For ReLATE, we use DINOv2-ViT-B/14 as the visual backbone. The model is trained with a batch size of 32 on NVIDIA A800 GPUs. The training image size is $224 \times 224$, and the test image size is $280 \times 280$. We use SGD optimizer with an initial learning rate of 0.03, momentum of 0.9, and weight decay of $5 \times 10^{-4}$. A cosine learning-rate scheduler is adopted for 160 epochs with 175 warm-up steps. We use $K_q{=}8$ internal learnable queries and
$N_{\mathrm{sup}}{=}2$ query-derived supervised branches.
The eight reliability-guided internal queries are linearly remapped
along the query dimension into two branch representations.
Together with the CLS-token branch and the GeM-pooled spatial branch,
they form
$N_h{=}N_{\mathrm{sup}}+2{=}4$
final descriptor branches per view, each supervised by a location
classifier. The training objective combines the resulting multi-branch location classification loss, a cross-view Multi-Similarity metric loss, and a cross-view prediction-consistency loss, as detailed in Section~\ref{sec:training_objective}. Unless otherwise specified, the same trained checkpoint is used for all corruption types and severity levels within each dataset.

\begin{figure}[tb]

\centering
	\includegraphics[width=\linewidth]{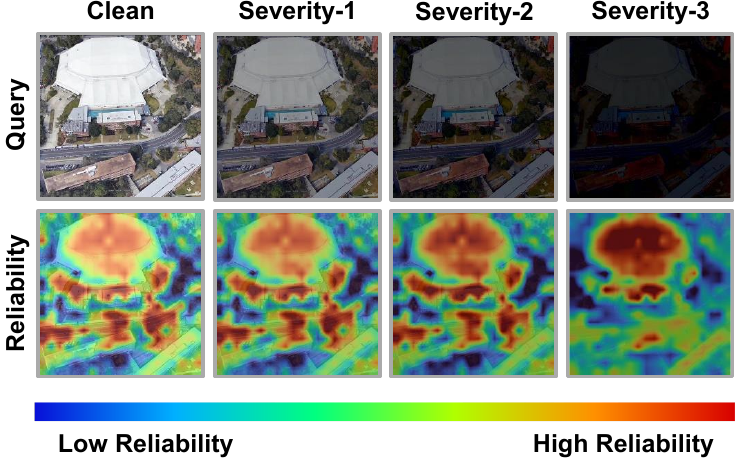}
	\caption{
		Visualization of the structure-smoothed reliability field
		$\widetilde{\mathbf{R}}^{v}$ in Eq.~\eqref{eq:rel_smooth} under night
		corruption of increasing severity. Top: the clean UAV query and its
		corrupted versions at severity levels 1, 2, and 3. Bottom: the
		corresponding reliability maps, where warmer colors indicate higher
		relative model scores within each displayed map. Spatial emphasis on
		the dominant scene structure remains visible even at severity 3, when
		much of the local texture is no longer visible. The colors illustrate
		spatial patterns and do not establish calibrated reliability differences
		between images. 
	}
	\label{fig:reliability_severities}

\end{figure}

\begin{figure}[tb]

\centering
	\includegraphics[width=\linewidth]{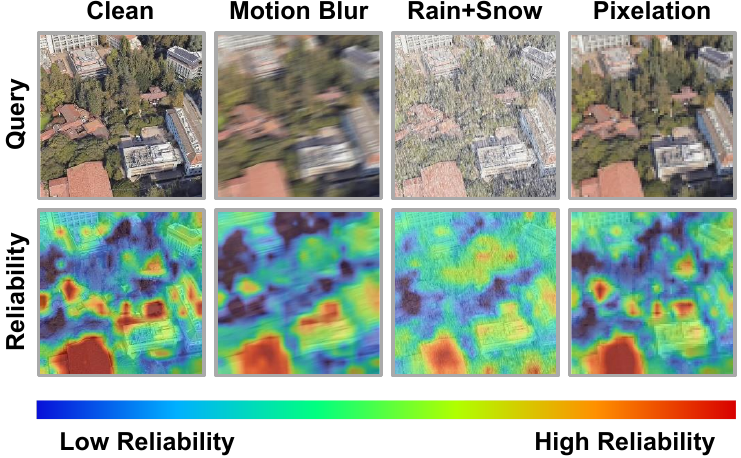}
	\caption{
		Visualization of the structure-smoothed reliability field
		$\widetilde{\mathbf{R}}^{v}$ in Eq.~\eqref{eq:rel_smooth} under different
		corruption types. Top: clean and corrupted UAV queries under motion blur,
		the compound rain+snow corruption, and pixelation. Bottom: the
		corresponding reliability maps, where warmer colors indicate higher
		relative model scores within each displayed map. The maps illustrate
		spatially varying responses around building contours and road layouts
		under the evaluated corruptions; their colors are not interpreted as
		calibrated confidence. All maps are produced by the same clean-trained model without
		corruption labels.
	}
	\label{fig:reliability_types}

\end{figure}

\subsection{Comparison with State-of-the-art Methods}
\label{sec:sota_comparison}

A central claim of ReLATE is that modeling evidence reliability matters most when observations are severely corrupted. Table~\ref{tab:severity_level_summary} examines this directly by grouping all corruption types by severity. Two patterns emerge. First, ReLATE improves over the strongest baseline QDFL at every severity level and in both retrieval directions, indicating that the gain is systematic rather than confined to a particular difficulty regime. Second, the advantage generally becomes more pronounced as degradation intensifies. On University-1652-Deg under Drone $\rightarrow$ Satellite, the R@1 margin widens consistently from $+2.40$ at severity 1 to $+4.90$ at severity 2 and $+5.75$ at severity 3, while ReLATE maintains positive gains at all severity levels on SUES-200-Deg. Fig.~\ref{fig:reliability_severities} provides a qualitative view of the learned spatial weighting under increasing night corruption. Across the illustrated severities, relatively high responses remain concentrated on the dominant building structure. These spatial patterns illustrate how the learned field can guide local evidence aggregation under degraded observations; the retrieval gains are established by the quantitative evaluation. 

To provide a more intuitive summary of performance consistency
across the large number of corrupted evaluation conditions,
Table~\ref{tab:rank_consistency_summary} reports a rank-based
robustness rating. Each dataset--direction--corruption combination
is treated as one condition, and methods receive decreasing scores
from 5 to 1 according to their condition-wise R@1 rank. On
University-1652-Deg and SUES-200-Deg, ReLATE obtains ratings of
4.65/5 and 4.67/5, respectively. It ranks first in 39 of the 54
conditions on each dataset, remains within the top two in 50 and
51 conditions, and never falls below third place in any of the
108 conditions. This result demonstrates that the advantage of
ReLATE is broadly distributed across datasets, retrieval directions,
and corruption types, rather than being driven by a small number
of favorable cases.

\begin{figure*}[tb]
	\centering
	\includegraphics[width=\textwidth]{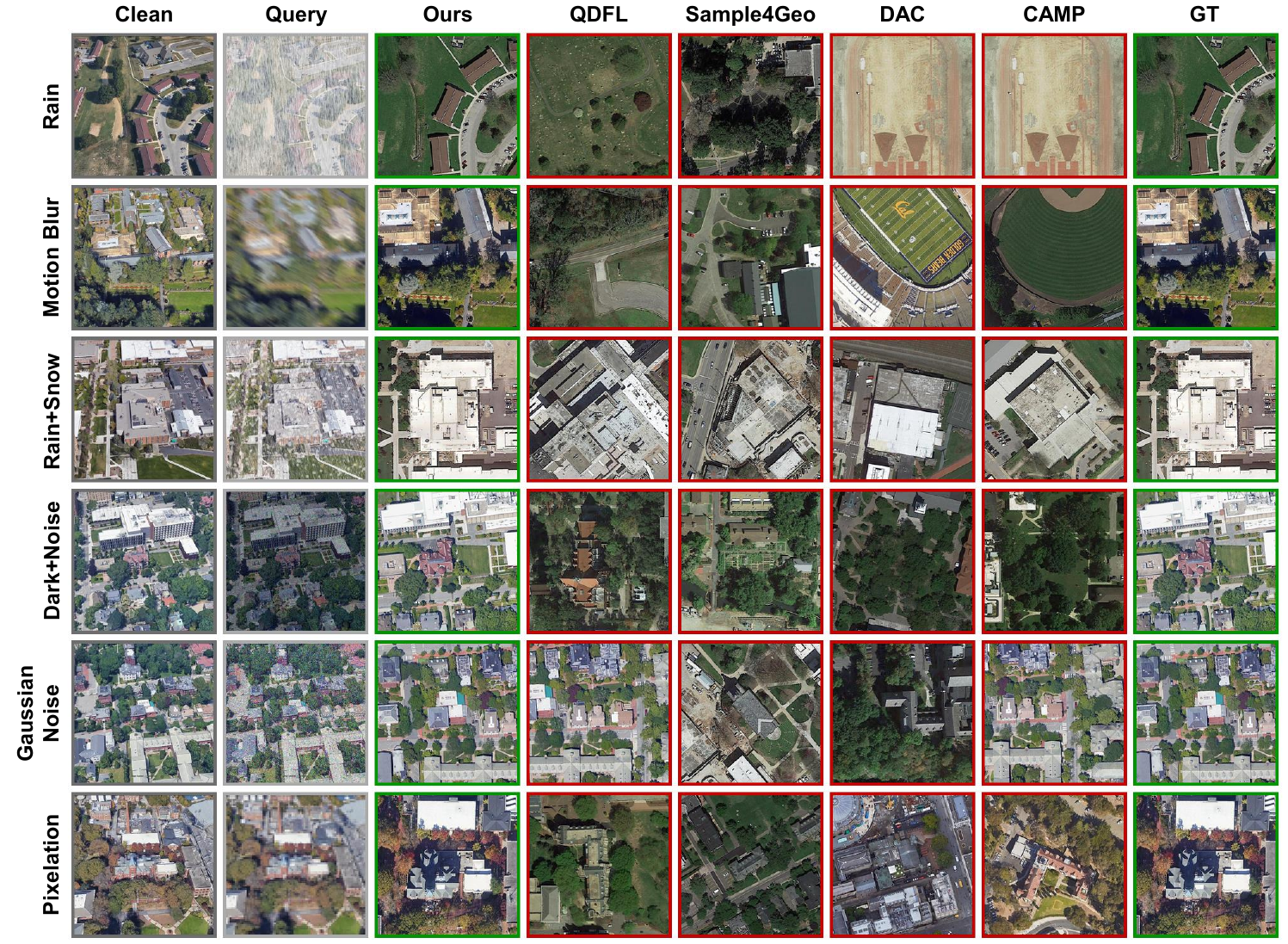}
	\caption{
		Qualitative retrieval results on University-1652-Deg under representative degraded UAV queries.
		Each row corresponds to one degraded UAV query, and each column shows the clean UAV image, the degraded query, the retrieved result of different methods, or the ground-truth satellite image. Green borders indicate correct matches, while red borders indicate incorrect matches. In these representative challenging examples, Ours retrieves
		the correct satellite image, whereas QDFL, Sample4Geo, DAC,
		and CAMP return incorrect matches.
	}
	\label{fig:u1652_qualitative}
\end{figure*}

\noindent\emph{1) Results on University-1652-Deg.}
The Drone $\rightarrow$ Satellite direction (Table~\ref{tab:u1652_drone_to_sat_all}) is
particularly challenging because the corrupted UAV observation directly
determines the query descriptor used to rank the clean satellite gallery. ReLATE attains the best corrupted average, raising R@1/AP from $65.39/68.55$ to $69.75/72.73$. Rather than enumerate individual cells, we note the shape of the improvement: the gain is broad-based, spanning appearance and visibility corruptions, such as rain, snow, nighttime, and brightness, as well as corruptions that erase local patterns, such as Gaussian and impulse noise, defocus and glass blur. These gains extend across multiple evaluated corruption families, supporting the usefulness of reliability-guided evidence learning beyond a single degradation category. The qualitative examples in Fig.~\ref{fig:u1652_qualitative} show the same effect at the instance level, where ReLATE recovers the correct match under queries whose discriminative regions are partially destroyed.
Fig.~\ref{fig:reliability_types} provides additional qualitative
evidence: under motion blur, rain+snow, and pixelation, the
relatively high responses remain concentrated around scene structures,
illustrating similar spatial emphasis across these different corruption types.

Compound corruptions, which superimpose multiple degradation factors, are the most diagnostic case for reliability modeling because they simultaneously reduce visibility and destroy local structure. Across the eight mixed and compound corruptions, ReLATE improves over QDFL by approximately $+5.78/+5.63$ percentage points in R@1/AP on the compound average. The advantage is not uniform. Under dark+rain+fog, ReLATE does not achieve the best R@1,
indicating that its advantage is not uniform across all
corruption types. Across the full compound panel, however, ReLATE is the most robust on average, showing improved average performance across the evaluated compound-corruption settings.

The corresponding Satellite $\rightarrow$ Drone results on
University-1652-Deg are reported in
Table~\ref{tab:u1652_sat_to_drone_all}. ReLATE increases the All-27 average from
$88.61/73.01$ achieved by QDFL to $89.81/75.20$ in R@1/AP, and obtains
the best results on most digital, noise, and blur corruptions. These
results show that reliability-guided evidence learning improves not only
query-side robustness in Drone $\rightarrow$ Satellite retrieval, but
also gallery-side robustness when degraded UAV observations must remain
aligned with a clean satellite query.

\noindent\emph{2) Results on SUES-200-Deg.}

\begin{figure*}[tb]
	\centering
	\includegraphics[
	width=0.95\textwidth,
	trim=5 5 5 5,
	clip
	]{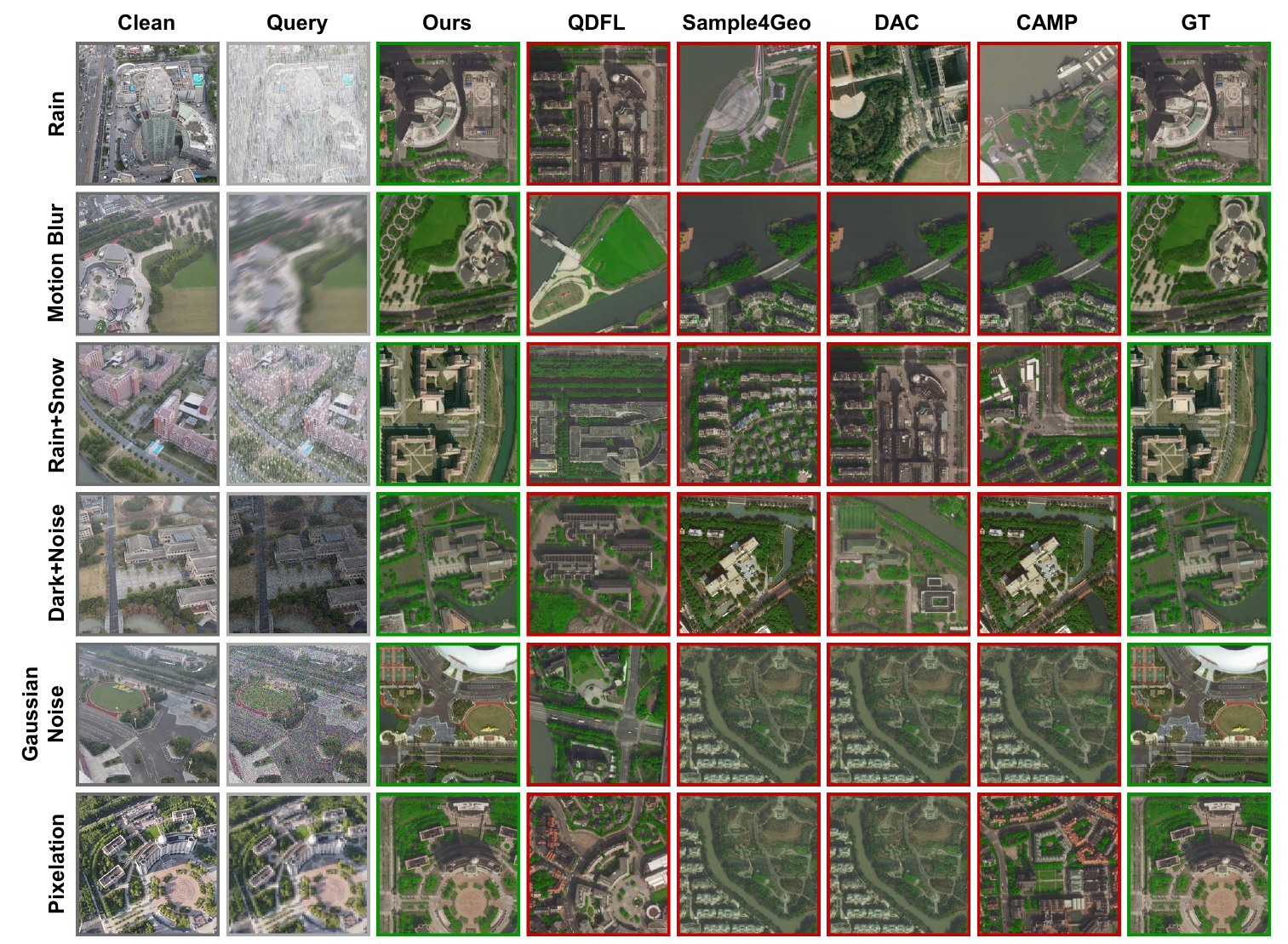}
	\caption{
		Qualitative retrieval results on SUES-200-Deg under representative degraded UAV queries.
		Each row corresponds to one degraded UAV query, and each column shows the clean UAV image,
		the degraded query, the retrieved result of different methods, or the ground-truth satellite image.
		Green borders indicate correct matches, while red borders indicate incorrect matches.
		In these representative challenging examples, Ours retrieves
		the correct satellite image, whereas QDFL, Sample4Geo, DAC,
		and CAMP return incorrect matches.
	}
	\label{fig:sues_qualitative}
\end{figure*}

SUES-200-Deg further introduces scale and viewpoint changes through multi-height UAV acquisition, and Tables~\ref{tab:sues_drone_to_sat_all} and~\ref{tab:sues_sat_to_drone_all} report dataset-level results averaged over all heights. In the Drone $\rightarrow$ Satellite setting, ReLATE achieves the best All-27 corrupted average, improving QDFL from $70.47/77.73$ to $73.78/80.45$ in R@1/AP. Clear gains are observed under pixelation, sensor noise, blur, and compound corruptions, with an average improvement of $+4.86/+4.24$ over QDFL on the eight compound types. These results support the intended role of reliability-guided learning in suppressing corrupted local evidence while preserving trustworthy spatial cues. The qualitative examples in Fig.~\ref{fig:sues_qualitative} show a consistent advantage under rain, rain+snow, motion blur, dark+noise, Gaussian noise, and pixelation.

In the Satellite $\rightarrow$ Drone setting, ReLATE also achieves the best All-27 average, improving QDFL from $88.67/77.39$ to $90.36/78.68$. The improvements include $+3.44/+2.87$ under pixelation, $+5.42/+4.41$ under motion blur, and $+5.62/+5.00$ under rain+motion. Although the overall margin is smaller than that in Drone $\rightarrow$ Satellite retrieval, the positive gains in both directions show that ReLATE remains effective whether degraded UAV observations appear on the query side or the gallery side.

\begin{table}[b]
	\centering
	\caption{
		Height-wise corrupted-test performance on SUES-200-Deg.
		Each entry is R@1/AP and is macro-averaged over all 27 corruption types
		and severity levels 1, 2, and 3.
		Clean subsets are excluded.
		Best and second-best results at each height are marked in
		\textcolor{red}{red} and \textcolor{blue}{blue}, respectively.
	}
	\label{tab:sues_heightwise}
	\fontsize{8pt}{9.5pt}\selectfont
	\setlength{\tabcolsep}{1.5pt}
	\renewcommand{\arraystretch}{1.05}
	
		\begin{tabular*}{\linewidth}{@{}ll@{\extracolsep{\fill}}cccc@{}}
			\toprule
			Direction & Method & H150 & H200 & H250 & H300 \\
			\midrule
			
			\multirow{2}{*}{D2S}
			& QDFL
			& \secondres{68.10}/\secondres{76.34}
			& \secondres{69.85}/\secondres{77.21}
			& \secondres{71.75}/\secondres{78.44}
			& \secondres{72.19}/\secondres{78.94} \\
			
			& \textbf{Ours}
			& \bestres{73.62}/\bestres{80.87}
			& \bestres{70.28}/\bestres{77.42}
			& \bestres{75.68}/\bestres{81.94}
			& \bestres{75.52}/\bestres{81.58} \\
			
			\midrule
			
			\multirow{2}{*}{S2D}
			& QDFL
			& \secondres{88.24}/\secondres{75.23}
			& \bestres{89.01}/\bestres{77.16}
			& \secondres{89.40}/\secondres{79.22}
			& \secondres{88.01}/\secondres{77.95} \\
			
			& \textbf{Ours}
			& \bestres{91.10}/\bestres{77.41}
			& \secondres{88.81}/\secondres{76.20}
			& \bestres{90.90}/\bestres{80.03}
			& \bestres{90.63}/\bestres{81.08} \\
			
			\bottomrule
		\end{tabular*}
\end{table}

Table~\ref{tab:sues_heightwise} reports the SUES-200-Deg results at
each flight height. Under D2S, ReLATE outperforms QDFL at all four
heights, with the largest gain at H150 ($+5.52/+4.53$ in R@1/AP). Under S2D, ReLATE improves the results at H150, H250, and
H300, whereas H200 is the only exception, with differences of
$-0.20/-0.96$. This variation shows that the benefit does not change
monotonically with altitude, but instead depends on the joint effects
of viewpoint, scale, degradation, and the amount of reliable evidence
that remains.

\begin{table}[b]
	\centering
	\caption{
		Zero-shot cross-dataset generalization from University-1652 to
		SUES-200-Deg at H300 under D2S. All models use only clean
		University-1652 training, without target-domain training or
		adaptation. Entries are R@1/AP averaged over three severity
		levels; All-27 excludes Clean.
	}
	\label{tab:cross_dataset_generalization}
	\fontsize{8pt}{9.5pt}\selectfont
	\setlength{\tabcolsep}{1.5pt}
	\renewcommand{\arraystretch}{1.05}
	
	\begin{tabular*}{\linewidth}{@{}l@{\extracolsep{\fill}}cccc@{}}
		\toprule
		Method & Clean & Core Avg. & Comp. Avg. & All-27 \\
		\midrule
		CAMP
		& 91.93/93.53
		& 56.09/60.00
		& 36.76/41.42
		& 50.36/54.49 \\
		
		QDFL
		& \bestres{96.58}/\bestres{97.96}
		& \secondres{74.57}/\secondres{80.89}
		& \secondres{58.07}/\secondres{66.49}
		& \secondres{69.68}/\secondres{76.62} \\
		
		\textbf{Ours}
		& \secondres{96.50}/\secondres{97.82}
		& \bestres{76.84}/\bestres{82.82}
		& \bestres{61.59}/\bestres{69.65}
		& \bestres{72.33}/\bestres{78.92} \\
		\bottomrule
	\end{tabular*}
\end{table}

\noindent\emph{3) Cross-Dataset Generalization.}
To further evaluate transferability beyond the training dataset,
Table~\ref{tab:cross_dataset_generalization} directly tests the
University-1652-trained models on SUES-200-Deg. ReLATE achieves
$72.33/78.92$ on All-27, outperforming QDFL by $+2.65/+2.30$ and CAMP
by $+21.97/+24.43$ in R@1/AP. Its gain over QDFL increases from
$+2.27/+1.93$ on core corruptions to $+3.52/+3.16$ on compound
corruptions, while clean performance remains comparable. These results
show that reliability-guided evidence learning transfers effectively to
an unseen target dataset and is especially beneficial under compound
degradation.

\section{Ablation Study}
\label{sec:ablation}

\subsection{Component-wise Ablation}
\label{subsec:component_ablation}

We evaluate SRE alone and a reliability-neutral counterpart of the
RATE structure. Since the full RATE module consumes the reliability
field produced by SRE, its reliability-adaptive form is evaluated on
top of SRE in the complete model. SRE learns a structure-smoothed reliability field to identify trustworthy local evidence, while RATE adaptively regulates how reliable token evidence contributes to the final descriptor. The full model, denoted as Full, combines both components.

Table~\ref{tab:ablation_component} jointly reports the component-
and mechanism-level ablation results. Compared with the Base model, adding SRE alone improves the corrupted average from 65.39/68.55 to 66.65/69.72 in terms of R@1/AP, bringing gains of +1.26/+1.17. This result shows that explicitly learning reliability cues is useful for degraded cross-view matching. Since corruptions often destroy local texture and weaken discriminative structures, SRE provides a structural reliability basis that helps the model distinguish more trustworthy regions from degraded visual responses.

Since RATE requires the reliability field estimated by SRE, it cannot be completely isolated once SRE is physically removed. For the corresponding ablation, we therefore retain the token evidence aggregation and descriptor injection structure of RATE, but replace the learned reliability field with a neutral zero field, thereby disabling spatially varying, reliability-dependent regulation while preserving the remaining computation. This variant, denoted as Base+RATE$^{\dagger}$, brings a more pronounced improvement: the corrupted average increases to 68.09/71.22, corresponding to gains of +2.70/+2.67 over the Base model, and the improvement is especially clear on compound corruptions, where the variant improves the Base model from 53.96/57.66 to 58.29/61.89, yielding gains of +4.33/+4.23. This result indicates that explicitly aggregating pooled local token evidence and injecting it into the query representations benefits descriptor construction even without learned spatial reliability, especially when multiple degradation factors are mixed and a globally aggregated descriptor alone becomes insufficient.

The full model achieves the best performance across all corrupted
settings, obtaining 69.75/72.73 on the corrupted average and improving
the Base model by +4.36/+4.18. More
importantly, adding RATE on top of SRE further improves the corrupted
average from 66.65/69.72 to 69.75/72.73, corresponding to gains of
+3.10/+3.01 in R@1/AP. This incremental improvement supports the effectiveness of
reliability-adaptive token evidence regulation.

\begin{table}[b]
	\centering
	\caption{
		Component- and mechanism-level ablation on University-1652-Deg
		under Drone $\rightarrow$ Satellite retrieval.
		Each entry is R@1/AP.
		Green values denote absolute gains over Base.
		Base+RATE$^{\dagger}$ retains the token evidence aggregation
		and injection structure of RATE with a neutral reliability field.
		The $\alpha=0$ variant disables structural smoothing, while
		Global-$\lambda$ replaces the input-dependent $\lambda^v$
		with one learnable scalar shared across all images and views.
	}
	\label{tab:ablation_component}
	\fontsize{8pt}{9.5pt}\selectfont
	\setlength{\tabcolsep}{1.5pt}
	\renewcommand{\arraystretch}{1.05}
	
	\begin{tabular*}{\linewidth}{@{\extracolsep{\fill}}lcccc@{}}
		\toprule
		Variant & Clean & Corr. Avg. & Core Avg. & Comp. Avg. \\
		\midrule
		
		Base
		& 95.00/95.83
		& 65.39/68.55
		& 70.21/73.13
		& 53.96/57.66 \\
		
		Base+SRE
		& \metricgain{95.31/96.20}{(+0.31/+0.37)}
		& \metricgain{66.65/69.72}{(+1.26/+1.17)}
		& \metricgain{71.38/74.22}{(+1.17/+1.09)}
		& \metricgain{55.44/59.05}{(+1.48/+1.39)} \\
		
		Base+RATE$^{\dagger}$
		& \metricgain{95.19/96.11}{(+0.19/+0.28)}
		& \metricgain{68.09/71.22}{(+2.70/+2.67)}
		& \metricgain{72.22/75.15}{(+2.01/+2.02)}
		& \metricgain{58.29/61.89}{(+4.33/+4.23)} \\
		
		\midrule
		
		$\alpha=0$
		& 94.40/95.33
		& 67.17/70.26
		& 70.84/73.73
		& 58.45/62.04 \\
		
		Global-$\lambda$
		& 94.24/95.18
		& 66.79/69.92
		& 71.27/74.18
		& 56.14/59.80 \\
		
		\textbf{Full}
		& \metricgain{\textbf{95.42/96.36}}{(+0.42/+0.53)}
		& \metricgain{\textbf{69.75/72.73}}{(+4.36/+4.18)}
		& \metricgain{\textbf{73.96/76.71}}{(+3.75/+3.58)}
		& \metricgain{\textbf{59.74/63.29}}{(+5.78/+5.63)} \\
		
		\bottomrule
	\end{tabular*}
\end{table}

The gain is larger on compound corruptions, where the full model reaches
59.74/63.29 and improves the Base model by
+5.78/+5.63. In particular, compared with Base+RATE$^{\dagger}$, the full model
further improves the corrupted average by $+1.66/+1.51$.
This shows that the neutral token aggregation-and-injection structure
alone does not account for the full gain, and that learned reliability
throughout the complete SRE--RATE pipeline provides additional benefit. SRE focuses on discovering reliable spatial evidence, while RATE further converts reliability into effective token-level evidence regulation. Their combination forms a complete reliability-guided matching mechanism, leading to stronger robustness than the SRE-only and reliability-neutral aggregation-and-injection variants.

It is also worth noting that the clean performance is not sacrificed. The full model achieves
95.42/96.36 on the clean subset, outperforming the Base model by
+0.42/+0.53. This indicates that the proposed components
do not simply trade clean-image discrimination for corrupted-image robustness.
Instead, they improve the representation by enhancing the use of reliable structural
evidence, which benefits both clean and degraded retrieval scenarios.

\subsection{Structural Smoothing and Adaptive Regulation}
\label{subsec:mechanism_ablation}

We further isolate two mechanism-level designs in ReLATE:
structural smoothing in SRE and input-dependent evidence regulation
in RATE. All variants use identical training and evaluation settings
and differ only in the designated mechanism. In the $\alpha=0$
variant, structural blending is disabled, such that the final
reliability field degenerates to the raw token-wise reliability field,
while reliability estimation, token modulation, token evidence
aggregation, and the remaining RATE operations are retained.
In the Global-$\lambda$ variant, the input-dependent coefficient
$\lambda^v$ is replaced by a single sigmoid-parameterized learnable
scalar shared across all images and views, while reliability estimation,
structural smoothing, reliability-weighted token aggregation, and
descriptor injection remain unchanged.

As reported in Table~\ref{tab:ablation_component}, disabling
structural smoothing reduces the corrupted average from
69.75/72.73 to 67.17/70.26, corresponding to decreases of
$2.58/2.47$ in R@1/AP. The full model further exceeds the
$\alpha=0$ variant by $3.12/2.98$ on core corruptions and
$1.29/1.25$ on compound corruptions. Since the remaining
reliability estimation and utilization operations are preserved,
this comparison isolates the contribution of incorporating local
spatial consistency into the reliability field. The results show
that raw token-wise reliability alone is insufficient, and that
the learnable structural blend provides a more stable basis for
reliability-guided representation learning.

For Global-$\lambda$, the shared regulation coefficient converges
to approximately $0.511$, showing that the control variant learns
a globally optimized evidence-injection strength rather than using
a manually fixed value. Nevertheless, the full input-dependent
regulation improves the corrupted average by $2.96/2.81$ over
Global-$\lambda$, with a larger improvement of $3.60/3.49$ on
compound corruptions. This result demonstrates that the benefit of
RATE cannot be explained by a fixed residual injection coefficient
alone. Instead, adapting the contribution of reliable token evidence
to the reliability state of each input is important, particularly when
compound degradations produce highly heterogeneous local evidence.
\subsection{Robustness Across Degradation Severities}
\label{subsec:severity_ablation}

\begin{table}[b]
	\centering
	\caption{
		Severity-wise component and mechanism ablation on
		University-1652-Deg under Drone $\rightarrow$ Satellite
		retrieval.
		Each entry is R@1/AP and is macro-averaged over all
		27 corruption types at the corresponding severity level.
		Green values, where shown, denote absolute gains over Base.
		The $\alpha=0$ variant disables structural smoothing,
		while Global-$\lambda$ uses one learnable regulation
		coefficient shared across all images and views.
	}
	\label{tab:ablation_component_severity}
	\fontsize{8pt}{9.5pt}\selectfont
	\setlength{\tabcolsep}{1.5pt}
	\renewcommand{\arraystretch}{1.05}
	\begin{tabular*}{\linewidth}{@{\extracolsep{\fill}}lccc@{}}
		\toprule
		Variant & Sev. 1 & Sev. 2 & Sev. 3 \\
		\midrule
		
		Base
		& 85.36/87.30
		& 66.33/69.77
		& 44.50/48.59 \\
		
		Base+SRE
		& \metricgain{86.66/88.48}{(+1.30/+1.18)}
		& \metricgain{68.04/71.35}{(+1.71/+1.58)}
		& \metricgain{45.27/49.34}{(+0.77/+0.75)} \\
		
		Base+RATE$^{\dagger}$
		& \metricgain{87.10/88.94}{(+1.74/+1.64)}
		& \metricgain{69.73/73.07}{(+3.40/+3.30)}
		& \metricgain{47.45/51.65}{(+2.95/+3.06)} \\
		
		$\alpha=0$
		& \metricgain{87.01/88.84}{(+1.65/+1.54)}
		& \metricgain{68.39/71.72}{(+2.06/+1.95)}
		& \metricgain{46.10/50.23}{(+1.60/+1.64)} \\
		
		Global-$\lambda$
		& \metricgain{86.46/88.34}{(+1.10/+1.04)}
		& \metricgain{68.16/71.54}{(+1.83/+1.77)}
		& \metricgain{45.75/49.88}{(+1.25/+1.29)} \\
		
		Full
		& \metricgain{87.76/89.48}{(+2.40/+2.18)}
		& \metricgain{71.23/74.43}{(+4.90/+4.66)}
		& \metricgain{50.25/54.29}{(+5.75/+5.70)} \\
		
		\bottomrule
	\end{tabular*}
\end{table}

Table~\ref{tab:ablation_component_severity} further examines the
component contributions across different degradation strengths.
Both Base+SRE and Base+RATE$^{\dagger}$ outperform the Base model
at all three severity levels, showing that structure-smoothed
reliability learning and token evidence aggregation and injection
remain beneficial across different degradation regimes.
More importantly, the gains of the full model over Base increase
from $+2.40/+2.18$ at severity 1 to $+4.90/+4.66$ at severity 2
and $+5.75/+5.70$ at severity 3.
This widening advantage indicates that combining learned reliability
with adaptive token evidence regulation becomes increasingly valuable
as degradation intensifies and trustworthy local evidence becomes
more limited.
At severity 3, the full model further exceeds Base+SRE by
$+4.98/+4.95$ and Base+RATE$^{\dagger}$ by $+2.80/+2.64$,
further supporting the complementary roles of reliability-guided
evidence learning and token evidence regulation under severe
corruptions.

The two mechanism-level variants provide further insight into why
the complete reliability pathway becomes more effective under stronger
degradation. Compared with $\alpha=0$, the full model improves
R@1/AP by $+0.75/+0.64$, $+2.84/+2.71$, and $+4.15/+4.06$
at severity levels 1, 2, and 3, respectively. Thus, the benefit of
structure-smoothed reliability increases substantially as degradation
intensifies, which is consistent with its role in reducing fragmented
and unstable token-wise reliability responses.

A similar trend is observed for input-dependent regulation.
Compared with Global-$\lambda$, the full model gains
$+1.30/+1.14$ at severity 1, $+3.07/+2.89$ at severity 2,
and $+4.50/+4.41$ at severity 3. The widening margin shows that
a single globally optimized injection strength is increasingly
insufficient under stronger corruptions, whereas the proposed
$\lambda^v$ can regulate reliable token evidence according to the
reliability state of each input. Together, these results support both the structure-smoothed reliability construction in SRE
and the input-dependent evidence regulation in RATE.

\section{Conclusion}
\label{sec:conclusion}

This paper investigates UAV-satellite geo-localization under degraded remote sensing observations. We construct UAVSat-Deg, including University-1652-Deg and SUES-200-Deg, to provide a systematic clean-training corrupted-testing benchmark with diverse degradation types, severity levels, bidirectional retrieval tasks, and multi-height UAV settings. Different from degradation-aware or restoration-based protocols, our benchmark evaluates image-only robustness without using corruption labels, text prompts, or auxiliary modalities, thereby exposing the robustness limitations of existing UAV-satellite geo-localization methods under degraded conditions.

To address this problem, we propose ReLATE, a reliability-guided framework that learns to emphasize trustworthy visual evidence and suppress unreliable token responses caused by weather, illumination, noise, blur, and compound corruptions. Extensive experiments show that ReLATE achieves stronger corrupted-test performance on both University-1652-Deg and SUES-200-Deg while maintaining competitive clean retrieval accuracy. Qualitative results and ablation studies further support that SRE and RATE are complementary for robust descriptor construction. In the future, we plan to extend UAVSat-Deg with real-captured degradations and satellite-side perturbations, and to further validate the proposed reliability layer on other representation substrates and localization pipelines. We hope UAVSat-Deg and ReLATE can serve as a useful benchmark and baseline for future research on robust UAV-satellite geo-localization in degraded environments.


\par\medskip
\noindent\textbf{Acknowledgements}\quad This work was supported by the National Natural Science Foundation of China under Grant Nos.~624B2051 (Special Program) and U25A6022.
\par\medskip
\noindent\textbf{Data availability}\quad The code and UAVSat-Deg benchmark will be made available at \url{https://github.com/JHC626/RELATE}.
\par\medskip
\noindent\textbf{Conflict of interest}\quad The authors declare that they have no conflict of interest.
\par


\medskip
\begin{thebibliography}{10}
\setlength{\baselineskip}{10pt}
\providecommand{\url}[1]{\texttt{#1}}
\providecommand{\urlprefix}{URL }
\providecommand{\bibinfo}[2]{#2}

\bibitem{zheng2020university}
\bibinfo{author}{Zheng Z}, \bibinfo{author}{Wei Y}, \bibinfo{author}{Yang Y}.
\newblock \bibinfo{title}{University-1652: A multi-view multi-source benchmark
  for drone-based geo-localization}.
\newblock In: \bibinfo{booktitle}{Proceedings of the 28th ACM international
  conference on Multimedia}, \bibinfo{year}{2020}.
\newblock \bibinfo{pages}{1395--1403}

\bibitem{zhu2023sues}
\bibinfo{author}{Zhu R}, \bibinfo{author}{Yin L}, \bibinfo{author}{Yang M},
  et~al.
\newblock \bibinfo{title}{Sues-200: A multi-height multi-scene cross-view image
  benchmark across drone and satellite}.
\newblock \bibinfo{journal}{IEEE Trans Circuits Syst Video Technol}, \bibinfo{year}{2023}, \bibinfo{volume}{33}:
  \bibinfo{pages}{4825--4839}

\bibitem{lin2015learning}
\bibinfo{author}{Lin T~Y}, \bibinfo{author}{Cui Y}, \bibinfo{author}{Belongie
  S}, et~al.
\newblock \bibinfo{title}{Learning deep representations for ground-to-aerial
  geolocalization}.
\newblock In: \bibinfo{booktitle}{Proceedings of the IEEE conference on
  computer vision and pattern recognition}, \bibinfo{year}{2015}.
\newblock \bibinfo{pages}{5007--5015}

\bibitem{hu2018cvm}
\bibinfo{author}{Hu S}, \bibinfo{author}{Feng M}, \bibinfo{author}{Nguyen R~M},
  et~al.
\newblock \bibinfo{title}{Cvm-net: Cross-view matching network for image-based
  ground-to-aerial geo-localization}.
\newblock In: \bibinfo{booktitle}{Proceedings of the IEEE conference on
  computer vision and pattern recognition}, \bibinfo{year}{2018}.
\newblock \bibinfo{pages}{7258--7267}

\bibitem{shi2019spatial}
\bibinfo{author}{Shi Y}, \bibinfo{author}{Liu L}, \bibinfo{author}{Yu X},
  et~al.
\newblock \bibinfo{title}{Spatial-aware feature aggregation for image based
  cross-view geo-localization}.
\newblock \bibinfo{journal}{Advances in Neural Information Processing Systems},
  \bibinfo{year}{2019}, \bibinfo{volume}{32}

\bibitem{yang2021cross}
\bibinfo{author}{Yang H}, \bibinfo{author}{Lu X}, \bibinfo{author}{Zhu Y}.
\newblock \bibinfo{title}{Cross-view geo-localization with layer-to-layer
  transformer}.
\newblock \bibinfo{journal}{Advances in Neural Information Processing Systems},
  \bibinfo{year}{2021}, \bibinfo{volume}{34}: \bibinfo{pages}{29009--29020}

\bibitem{zhu2022transgeo}
\bibinfo{author}{Zhu S}, \bibinfo{author}{Shah M}, \bibinfo{author}{Chen C}.
\newblock \bibinfo{title}{Transgeo: Transformer is all you need for cross-view
  image geo-localization}.
\newblock In: \bibinfo{booktitle}{Proceedings of the IEEE/CVF Conference on
  Computer Vision and Pattern Recognition}, \bibinfo{year}{2022}.
\newblock \bibinfo{pages}{1162--1171}

\bibitem{deuser2023sample4geo}
\bibinfo{author}{Deuser F}, \bibinfo{author}{Habel K}, \bibinfo{author}{Oswald
  N}.
\newblock \bibinfo{title}{Sample4geo: Hard negative sampling for cross-view
  geo-localisation}.
\newblock In: \bibinfo{booktitle}{Proceedings of the IEEE/CVF International
  Conference on Computer Vision}, \bibinfo{year}{2023}.
\newblock \bibinfo{pages}{16847--16856}

\bibitem{dosovitskiy2020image}
\bibinfo{author}{Dosovitskiy A}, \bibinfo{author}{Beyer L},
  \bibinfo{author}{Kolesnikov A}, et~al.
\newblock \bibinfo{title}{An image is worth 16x16 words: Transformers for image
  recognition at scale}.
\newblock \bibinfo{journal}{arXiv preprint arXiv:2010.11929},
  \bibinfo{year}{2020}

\bibitem{oquab2023dinov2}
\bibinfo{author}{Oquab M}, \bibinfo{author}{Darcet T},
  \bibinfo{author}{Moutakanni T}, et~al.
\newblock \bibinfo{title}{Dinov2: Learning robust visual features without
  supervision}.
\newblock \bibinfo{journal}{arXiv preprint arXiv:2304.07193},
  \bibinfo{year}{2023}

\bibitem{zhang2024benchmarking}
\bibinfo{author}{Zhang Q}, \bibinfo{author}{Zhu Y}.
\newblock \bibinfo{title}{Benchmarking the robustness of cross-view
  geo-localization models}.
\newblock In: \bibinfo{booktitle}{European Conference on Computer Vision},
  \bibinfo{organization}{Springer}. \bibinfo{year}{2024}.
\newblock \bibinfo{pages}{36--53}

\bibitem{wang2024multiple}
\bibinfo{author}{Wang T}, \bibinfo{author}{Zheng Z}, \bibinfo{author}{Sun Y},
  et~al.
\newblock \bibinfo{title}{Multiple-environment self-adaptive network for
  aerial-view geo-localization}.
\newblock \bibinfo{journal}{Pattern Recognit}, \bibinfo{year}{2024},
  \bibinfo{volume}{152}: \bibinfo{pages}{110363}

\bibitem{feng2024multi}
\bibinfo{author}{Feng T}, \bibinfo{author}{Li Q}, \bibinfo{author}{Wang X},
  et~al.
\newblock \bibinfo{title}{Multi-weather cross-view geo-localization using
  denoising diffusion models}.
\newblock In: \bibinfo{booktitle}{Proceedings of the 2nd Workshop on UAVs in
  Multimedia: Capturing the World from a New Perspective},
  \bibinfo{year}{2024}.
\newblock \bibinfo{pages}{35--39}

\bibitem{wen2026weatherprompt}
\bibinfo{author}{Wen J}, \bibinfo{author}{Yu H}, \bibinfo{author}{Zheng Z}.
\newblock \bibinfo{title}{Weatherprompt: Multi-modality representation learning
  for all-weather drone visual geo-localization}.
\newblock \bibinfo{journal}{Advances in Neural Information Processing Systems},
  \bibinfo{year}{2026}, \bibinfo{volume}{38}: \bibinfo{pages}{32974--32994}

\bibitem{workman2015wide}
\bibinfo{author}{Workman S}, \bibinfo{author}{Souvenir R},
  \bibinfo{author}{Jacobs N}.
\newblock \bibinfo{title}{Wide-area image geolocalization with aerial reference
  imagery}.
\newblock In: \bibinfo{booktitle}{Proceedings of the IEEE International
  Conference on Computer Vision}, \bibinfo{year}{2015}.
\newblock \bibinfo{pages}{3961--3969}

\bibitem{wang2021each}
\bibinfo{author}{Wang T}, \bibinfo{author}{Zheng Z}, \bibinfo{author}{Yan C},
  et~al.
\newblock \bibinfo{title}{Each part matters: Local patterns facilitate
  cross-view geo-localization}.
\newblock \bibinfo{journal}{IEEE Trans Circuits Syst Video Technol}, \bibinfo{year}{2021}, \bibinfo{volume}{32}:
  \bibinfo{pages}{867--879}

\bibitem{dai2021transformer}
\bibinfo{author}{Dai M}, \bibinfo{author}{Hu J}, \bibinfo{author}{Zhuang J},
  et~al.
\newblock \bibinfo{title}{A transformer-based feature segmentation and region
  alignment method for uav-view geo-localization}.
\newblock \bibinfo{journal}{IEEE Trans Circuits Syst Video Technol}, \bibinfo{year}{2021}, \bibinfo{volume}{32}:
  \bibinfo{pages}{4376--4389}

\bibitem{ding2020practical}
\bibinfo{author}{Ding L}, \bibinfo{author}{Zhou J}, \bibinfo{author}{Meng L},
  et~al.
\newblock \bibinfo{title}{A practical cross-view image matching method between
  uav and satellite for uav-based geo-localization}.
\newblock \bibinfo{journal}{Remote Sens}, \bibinfo{year}{2020},
  \bibinfo{volume}{13}: \bibinfo{pages}{47}

\bibitem{tian2021uav}
\bibinfo{author}{Tian X}, \bibinfo{author}{Shao J}, \bibinfo{author}{Ouyang D},
  et~al.
\newblock \bibinfo{title}{Uav-satellite view synthesis for cross-view
  geo-localization}.
\newblock \bibinfo{journal}{IEEE Trans Circuits Syst Video Technol}, \bibinfo{year}{2021}, \bibinfo{volume}{32}:
  \bibinfo{pages}{4804--4815}

\bibitem{lin2022joint}
\bibinfo{author}{Lin J}, \bibinfo{author}{Zheng Z}, \bibinfo{author}{Zhong Z},
  et~al.
\newblock \bibinfo{title}{Joint representation learning and keypoint detection
  for cross-view geo-localization}.
\newblock \bibinfo{journal}{IEEE Trans Image Process},
  \bibinfo{year}{2022}, \bibinfo{volume}{31}: \bibinfo{pages}{3780--3792}

\bibitem{shen2023mccg}
\bibinfo{author}{Shen T}, \bibinfo{author}{Wei Y}, \bibinfo{author}{Kang L},
  et~al.
\newblock \bibinfo{title}{Mccg: A convnext-based multiple-classifier method for
  cross-view geo-localization}.
\newblock \bibinfo{journal}{IEEE Trans Circuits Syst Video Technol}, \bibinfo{year}{2023}, \bibinfo{volume}{34}:
  \bibinfo{pages}{1456--1468}

\bibitem{xia2024enhancing}
\bibinfo{author}{Xia P}, \bibinfo{author}{Wan Y}, \bibinfo{author}{Zheng Z},
  et~al.
\newblock \bibinfo{title}{Enhancing cross-view geo-localization with domain
  alignment and scene consistency}.
\newblock \bibinfo{journal}{IEEE Trans Circuits Syst Video Technol}, \bibinfo{year}{2024}, \bibinfo{volume}{34}:
  \bibinfo{pages}{13271--13281}

\bibitem{wu2024camp}
\bibinfo{author}{Wu Q}, \bibinfo{author}{Wan Y}, \bibinfo{author}{Zheng Z},
  et~al.
\newblock \bibinfo{title}{Camp: A cross-view geo-localization method using
  contrastive attributes mining and position-aware partitioning}.
\newblock \bibinfo{journal}{IEEE Trans Geosci Remote Sens}, \bibinfo{year}{2024}, \bibinfo{volume}{62}: \bibinfo{pages}{1--14}

\bibitem{lin2024self}
\bibinfo{author}{Lin J}, \bibinfo{author}{Luo Z}, \bibinfo{author}{Lin D},
  et~al.
\newblock \bibinfo{title}{A self-adaptive feature extraction method for
  aerial-view geo-localization}.
\newblock \bibinfo{journal}{IEEE Trans Image Process},
  \bibinfo{year}{2024}, \bibinfo{volume}{34}: \bibinfo{pages}{126--139}

\bibitem{du2024ccr}
\bibinfo{author}{Du H}, \bibinfo{author}{He J}, \bibinfo{author}{Zhao Y}.
\newblock \bibinfo{title}{Ccr: A counterfactual causal reasoning-based method
  for cross-view geo-localization}.
\newblock \bibinfo{journal}{IEEE Trans Circuits Syst Video Technol}, \bibinfo{year}{2024}, \bibinfo{volume}{34}:
  \bibinfo{pages}{11630--11643}

\bibitem{hu2025query}
\bibinfo{author}{Hu S}, \bibinfo{author}{Shi Z}, \bibinfo{author}{Jin T},
  et~al.
\newblock \bibinfo{title}{Query-driven feature learning for cross-view
  geo-localization}.
\newblock \bibinfo{journal}{IEEE Trans Geosci Remote Sens}, \bibinfo{year}{2025}

\bibitem{rodrigues2022global}
\bibinfo{author}{Rodrigues R}, \bibinfo{author}{Tani M}.
\newblock \bibinfo{title}{Global assists local: Effective aerial
  representations for field of view constrained image geo-localization}.
\newblock In: \bibinfo{booktitle}{Proceedings of the IEEE/CVF Winter Conference
  on Applications of Computer Vision}, \bibinfo{year}{2022}.
\newblock \bibinfo{pages}{3871--3879}

\bibitem{sun2023f3}
\bibinfo{author}{Sun B}, \bibinfo{author}{Liu G}, \bibinfo{author}{Yuan Y}.
\newblock \bibinfo{title}{F3-net: Multiview scene matching for drone-based
  geo-localization}.
\newblock \bibinfo{journal}{IEEE Trans Geosci Remote Sens}, \bibinfo{year}{2023}, \bibinfo{volume}{61}: \bibinfo{pages}{1--11}

\bibitem{wang2024learning}
\bibinfo{author}{Wang T}, \bibinfo{author}{Zheng Z}, \bibinfo{author}{Zhu Z},
  et~al.
\newblock \bibinfo{title}{Learning cross-view geo-localization embeddings via
  dynamic weighted decorrelation regularization}.
\newblock \bibinfo{journal}{IEEE Trans Geosci Remote Sens}, \bibinfo{year}{2024}, \bibinfo{volume}{62}: \bibinfo{pages}{1--12}

\bibitem{chen2024sdpl}
\bibinfo{author}{Chen Q}, \bibinfo{author}{Wang T}, \bibinfo{author}{Yang Z},
  et~al.
\newblock \bibinfo{title}{Sdpl: Shifting-dense partition learning for uav-view
  geo-localization}.
\newblock \bibinfo{journal}{IEEE Trans Circuits Syst Video Technol}, \bibinfo{year}{2024}, \bibinfo{volume}{34}:
  \bibinfo{pages}{11810--11824}

\bibitem{zhao2024transfg}
\bibinfo{author}{Zhao H}, \bibinfo{author}{Ren K}, \bibinfo{author}{Yue T},
  et~al.
\newblock \bibinfo{title}{Transfg: A cross-view geo-localization of satellite
  and uavs imagery pipeline using transformer-based feature aggregation and
  gradient guidance}.
\newblock \bibinfo{journal}{IEEE Trans Geosci Remote Sens}, \bibinfo{year}{2024}, \bibinfo{volume}{62}: \bibinfo{pages}{1--12}

\bibitem{ju2025video2bev}
\bibinfo{author}{Ju H}, \bibinfo{author}{Huang S}, \bibinfo{author}{Liu S},
  et~al.
\newblock \bibinfo{title}{Video2bev: Transforming drone videos to bevs for
  video-based geo-localization}.
\newblock In: \bibinfo{booktitle}{Proceedings of the IEEE/CVF International
  Conference on Computer Vision}, \bibinfo{year}{2025}.
\newblock \bibinfo{pages}{27073--27083}

\bibitem{chu2024towards}
\bibinfo{author}{Chu M}, \bibinfo{author}{Zheng Z}, \bibinfo{author}{Ji W},
  et~al.
\newblock \bibinfo{title}{Towards natural language-guided drones: Geotext-1652
  benchmark with spatial relation matching}.
\newblock In: \bibinfo{booktitle}{European Conference on Computer Vision},
  \bibinfo{organization}{Springer}. \bibinfo{year}{2024}.
\newblock \bibinfo{pages}{213--231}

\bibitem{zhao2025p2fcn}
\bibinfo{author}{Zhao Q}, \bibinfo{author}{Zhou J}, \bibinfo{author}{Wang T},
  et~al.
\newblock \bibinfo{title}{P2fcn: Environment-independent uav-view
  geo-localization via pixel-to-feature co-enhancement}.
\newblock \bibinfo{journal}{IEEE Trans Geosci Remote Sens}, \bibinfo{year}{2025}, \bibinfo{volume}{63}: \bibinfo{pages}{1--12}

\bibitem{hendrycks2019benchmarking}
\bibinfo{author}{Hendrycks D}, \bibinfo{author}{Dietterich T}.
\newblock \bibinfo{title}{Benchmarking neural network robustness to common
  corruptions and perturbations}.
\newblock \bibinfo{journal}{arXiv preprint arXiv:1903.12261},
  \bibinfo{year}{2019}

\bibitem{radford2021learning}
\bibinfo{author}{Radford A}, \bibinfo{author}{Kim J~W},
  \bibinfo{author}{Hallacy C}, et~al.
\newblock \bibinfo{title}{Learning transferable visual models from natural
  language supervision}.
\newblock In: \bibinfo{booktitle}{International conference on machine
  learning}, \bibinfo{organization}{PmLR}. \bibinfo{year}{2021}.
\newblock \bibinfo{pages}{8748--8763}

\bibitem{he2016deep}
\bibinfo{author}{He K}, \bibinfo{author}{Zhang X}, \bibinfo{author}{Ren S},
  et~al.
\newblock \bibinfo{title}{Deep residual learning for image recognition}.
\newblock In: \bibinfo{booktitle}{Proceedings of the IEEE conference on
  computer vision and pattern recognition}, \bibinfo{year}{2016}.
\newblock \bibinfo{pages}{770--778}

\bibitem{huang2017densely}
\bibinfo{author}{Huang G}, \bibinfo{author}{Liu Z}, \bibinfo{author}{Van
  Der~Maaten L}, et~al.
\newblock \bibinfo{title}{Densely connected convolutional networks}.
\newblock In: \bibinfo{booktitle}{Proceedings of the IEEE conference on
  computer vision and pattern recognition}, \bibinfo{year}{2017}.
\newblock \bibinfo{pages}{4700--4708}

\bibitem{pan2018two}
\bibinfo{author}{Pan X}, \bibinfo{author}{Luo P}, \bibinfo{author}{Shi J},
  et~al.
\newblock \bibinfo{title}{Two at once: Enhancing learning and generalization
  capacities via ibn-net}.
\newblock In: \bibinfo{booktitle}{Proceedings of the european conference on
  computer vision (ECCV)}, \bibinfo{year}{2018}.
\newblock \bibinfo{pages}{464--479}

\bibitem{liu2021swin}
\bibinfo{author}{Liu Z}, \bibinfo{author}{Lin Y}, \bibinfo{author}{Cao Y},
  et~al.
\newblock \bibinfo{title}{Swin transformer: Hierarchical vision transformer
  using shifted windows}.
\newblock In: \bibinfo{booktitle}{Proceedings of the IEEE/CVF international
  conference on computer vision}, \bibinfo{year}{2021}.
\newblock \bibinfo{pages}{10012--10022}

\bibitem{vaswani2017attention}
\bibinfo{author}{Vaswani A}, \bibinfo{author}{Shazeer N},
  \bibinfo{author}{Parmar N}, et~al.
\newblock \bibinfo{title}{Attention is all you need}.
\newblock \bibinfo{journal}{Advances in neural information processing systems},
  \bibinfo{year}{2017}, \bibinfo{volume}{30}

\bibitem{wang2019multi}
\bibinfo{author}{Wang X}, \bibinfo{author}{Han X}, \bibinfo{author}{Huang W},
  et~al.
\newblock \bibinfo{title}{Multi-similarity loss with general pair weighting for
  deep metric learning}.
\newblock In: \bibinfo{booktitle}{Proceedings of the IEEE/CVF conference on
  computer vision and pattern recognition}, \bibinfo{year}{2019}.
\newblock \bibinfo{pages}{5022--5030}

\bibitem{chen2025multi}
\bibinfo{author}{Chen Z}, \bibinfo{author}{Yang Z~X}, \bibinfo{author}{Rong
  H~J}.
\newblock \bibinfo{title}{Multi-level embedding and alignment network with
  consistency and invariance learning for cross-view geo-localization}.
\newblock \bibinfo{journal}{IEEE Trans Geosci Remote Sens}, \bibinfo{year}{2025}

\end{thebibliography}
\end{document}